\documentclass[11pt,oneside,a4paper]{article}

\usepackage{lineno}
\usepackage{authblk}

\modulolinenumbers[5]

\usepackage{amssymb}

\usepackage{amsmath,amssymb,amsfonts}
\usepackage{textcomp}
\usepackage{xcolor}
\usepackage{makecell}
\usepackage{lineno}
\usepackage{microtype}
\usepackage[utf8]{inputenc}

\usepackage{algorithm}
\usepackage[noend]{algpseudocode}
\usepackage{tikz}

\usepackage{xcolor} 
\definecolor{kamcol}{HTML}{007C64}

\usepackage{verbatim}

\usepackage{hyperref}
\hypersetup{
    colorlinks=true,
    linkcolor=kamcol,
    citecolor=kamcol,
    filecolor=magenta,      
    urlcolor=cyan,
}

\usepackage{chngpage}
\usepackage{subcaption}
\usepackage[font=small,labelfont=bf, justification=justified, format=plain]{caption}
\usepackage{cleveref}
\usepackage{afterpage}

\newcommand{\bx}{{\boldsymbol x}}

\newcommand{\bX}{{\boldsymbol X}}

\newcommand{\bI}{{\bf I}}
\newcommand{\knn}{k\textnormal{-nn}}
\newcommand{\kann}{k\textnormal{-}\widetilde{\operatorname{nn}}}

\newcommand{\LineFor}[2]{ \State \algorithmicfor\ {#1}\ \algorithmicdo\ {#2}}
\newcommand{\indi}[1]{\mathbf{1} \lbrace #1 \rbrace}

\DeclareMathOperator{\clust}{c}

\begin{document}

\title{A Distributed and Approximated Nearest Neighbors Algorithm for an Efficient Large Scale Mean Shift Clustering }

\author[1]{Ga\"el Beck}
\author[1]{Tarn Duong }
\author[1]{Mustapha Lebbah }
\author[1]{Hanane Azzag }
\author[1]{Christophe Cérin}

\affil[1]{Computer Science Laboratory of Paris North (LIPN, CNRS UMR 7030), University of Paris 13, F-93430 Villetaneuse France}

\date{}
\maketitle

\begin{abstract}
In this paper we target the class of modal clustering methods where clusters are defined in terms of the local modes of the probability density function which generates the data. The most well-known modal clustering method is the $k$-means clustering. Mean Shift clustering is a generalization of the $k$-means clustering which computes arbitrarily shaped clusters as defined as the basins of attraction to the local modes created by the density gradient ascent paths. Despite its potential, the Mean Shift approach is a computationally expensive method for unsupervised learning.
Thus, we introduce two contributions aiming to provide clustering algorithms with a linear time complexity, as opposed to the quadratic time complexity for the exact Mean Shift clustering. Firstly  we propose a scalable procedure to approximate the density gradient ascent. Second, our proposed scalable cluster labeling technique is presented. Both propositions are based on Locality Sensitive Hashing (LSH) to approximate nearest neighbors. These two techniques may be used for moderate sized datasets.  Furthermore, we show that using our  proposed approximations of the density gradient ascent as a pre-processing step in other clustering methods can also improve dedicated classification metrics.
For the latter, a distributed implementation, written for the Spark/Scala ecosystem is proposed.
For all these considered clustering methods, we present experimental results illustrating their labeling accuracy and their potential to solve concrete problems.
\end{abstract}

\providecommand{\keywords}[1]{\textbf{\textit{Index terms---}} #1}


\section{Introduction}
The goal of clustering or unsupervised learning is to assign cluster membership to unlabeled candidate points where the number and location of these clusters are unknown.
Current Mean Shift clustering \cite{fukunaga1973,fukunaga1975,Cheng:1995:MSM:628321.628711} algorithms contains computational bottlenecks with both kernel and nearest neighbor approaches: the former is due to the exact evaluation
of the kernel function, and the latter due to the exact nearest neighbor searches. We propose a new algorithm which resolves the computational inefficiencies of the nearest neighbor Mean Shift by using Locality Sensitive Hashing (LSH) \cite{indyk1998, datar2004, slaney2008} for approximate nearest neighbor searches to replace the exact nearest neighbor calculations in the density gradient ascent and in the cluster labeling stages. Compared to kernel approaches to Mean Shift clustering, which are $O(n^2)$ where $n$ is the size of the dataset, our nearest neighbors approach enables a scalable implementation of the gradient ascent and cluster labeling which are both $O(n)$. 

Furthermore, existing programming paradigms for dealing with parallelism, such as  MapReduce \cite{Dean:2008:MSD:1327452.1327492} and Message Passing Interface (MPI) \cite{Forum:1994:MMI:898758}, have been demonstrated the best practical choices for implementing these clustering  algorithms. MapReduce paradigm becomes popular and suited for data already stored on a distributed file system, which offers data replication, as well as the ability to execute computations, locally on each data node.
Thus, we implement this
approximate nearest neighbour Mean shift clustering algorithm on a distributed Apache Spark/Scala framework \cite{spark2012}, which
 allows us to carry out clustering on Big datasets. 
 

The organization of the paper is as follows. We review related works in Section~\ref{S:0}. In Section~\ref{S:1} we briefly introduce the problem prior to elaborating our algorithmic contributions for the gradient ascent and cluster labeling. Section~\ref{S:2} is related to the experiments conducted on the Grid'5000 testbed where we examine the role of the key tuning parameters for both accelerating the respond time and measuring the quality of the answer
Concluding remarks follow in Section~\ref{S:4}.

\section{Related works}
\label{S:0}


For comprehensive reviews of clustering, see for example the monographs as in \cite{bishop:2006:PRML,Aggarwal:2013:DCA:2535015}. For our purposes, we focus on the principal scalable clustering algorithm, i.e. on the 
well-known $k$-means algorithm \cite{DBLP:journals/corr/abs-1203-6402}. This algorithm has the advantage of having a single parameter $k$ which stands for the number of desired clusters. The algorithm works by moving $k$ prototypes towards the centroid formed by their closest data points. This process is repeated iteratively until the intra-class variance 
(the sum of squared distances from each data point within a cluster to its corresponding prototype) is minimized.
DBScan \cite{dbscanDistributed} is a well-known density based algorithm. It takes two parameters, $\varepsilon$ which defines the radius of the hypersphere and $minPts$ which is the minimum number of points above which a corresponding hypersphere is considered to be sufficiently dense. Each time the density threshold is reached, the data points in the same hypersphere belong to the same cluster, and the process is extended to include more data points until the data density falls under the threshold  determined by $\varepsilon$ and $minPts$. The remaining data points are considered to be noise. One notable advantage of this algorithm is its ability to  automatically detect the number of clusters with arbitrary shape. But it remains difficult to tune it correctly.
Like DBScan,  Mean Shift is a density based algorithm and can detect automatically the number of clusters with arbitrary shape. 
Most  studies on the Mean Shift clustering have focused on the kernel versions, e.g. \cite{Vedaldi2008,wu2007,2014-01-0170-Schugk}. The latter authors compared Gaussian, Cauchy and generalized Epanechnikov kernels to study the behaviour of tuning parameters of Mean Shift clustering.  

\section{Mean Shift clustering}
\label{S:1}
The Mean Shift algorithm and its variants consist in two major steps as described in Algorithm \ref{alg:meanshift}. The first important step (the density gradient ascent) is generally the most computationally intensive. This gradient ascent can be computed in different ways, such as with  kernel functions or as we propose in this paper,  nearest neighbors. The second step is the cluster labeling phase where we use the result from the first step to assign cluster labels to the original data points.

\begin{algorithm}[H]
\caption{Mean Shift principle}
\label{alg:meanshift}
\begin{algorithmic}[1]
\Statex {\bf Input:} points $\lbrace \bx_1, \dots, \bx_m \rbrace,$
\Statex {\bf Output:} cluster label $\{ \tilde{\clust}(\bx_1), \dots, \tilde{\clust}(\bx_m)\}$
\Statex { \bf Step 1:} Density gradient ascent; 
\Statex {\bf Step 2:} Cluster labeling;
\end{algorithmic}
\end{algorithm}

\subsection{Density gradient ascent}
\label{sub:ga}

The Mean Shift method for a $d$-dimensional point $\bx$, generates a sequence of points $\lbrace \bx_0, \bx_1, \dots \rbrace$ which follows the  gradient density ascent paths using the recurrence relation
\begin{equation}
\bx_{j+1} =  
\frac{1}{k} \sum_{\bX_i \in \knn(\bx_j)} \bX_i 
\label{eq:nnms}
\end{equation}
where $\bX_1, \dots, \bX_n$ is a random sample drawn from a common density $f$ and the $k$ nearest neighbors of $\bx$ are $\knn(\bx) = \{\bX_i : \lVert \bx - \bX_i \lVert \leq \delta_{(k)}(\bx) \}$ with $\delta_{(k)}(\bx)$ is the $k$-th nearest neighbor distance, and $\bx_0 = \bx$.  Eq.~\eqref{eq:nnms} gives the Mean Shift method its name since the current iterate $\bx_j$ is shifted to the sample mean of its $k$ nearest neighbors for the next iterate $\bx_{j+1}$.  
The gradient ascent paths towards the local modes produced by Eq.~\eqref{eq:nnms} form the basis of Algorithm~\ref{alg:nnga}, our nearest neighbor Mean Shift gradient ascent (NNGA).  

\indent The inputs to the NNGA are the data sample $\bX_1, \dots, \bX_n$ and the candidate points $\bx_1, \dots, \bx_m$, which we want to cluster (these can be the same as $\bX_1, \dots, \bX_n$, but this is not required); and the tuning parameters: the number of nearest neighbors $k$, the tolerance under which subsequent iterations in the Mean Shift update are considered to be convergent $\varepsilon_1$, the maximum number of iterations $j_{\max}$.


\begin{algorithm}[H]
\caption{NNGA -- Nearest Neighbor Gradient Ascent with exact $k$-nn}
\label{alg:nnga}
\begin{algorithmic}[1]
\Statex {\bf Input:} $\lbrace \bX_1, \dots, \bX_n \rbrace, \lbrace \bx_1, \dots, \bx_m \rbrace, k, \varepsilon_1, j_{\max}$ 
\Statex {\bf Output:} $\lbrace \bx_1^*, \dots, \bx_m^*\rbrace$
\State Compute similarity matrix and sort each row; 
\For{$\ell := 1$ to $m$}
\State $j := 0$; $\bx_{\ell,0} :=\bx_\ell$; 
\Statex /* Search for $k$-$nn$ based on similarity matrix */
\State $\bx_{\ell,1} := $ mean of $\knn (\bx_{\ell,0})$; 
\While{$\lVert \bx_{\ell, j+1}, \bx_{\ell,j} \lVert \, > \varepsilon_1$ {\bf or}  
$j < j_{\mathrm{max}}$}
\State $j := j+ 1$; 
\State $\bx_{\ell,j+1} := $ mean of $\knn(\bx_{\ell,j})$;
\EndWhile
\State{$\bx_\ell^* := \bx_{\ell,j}$;}
\EndFor
\end{algorithmic}
\end{algorithm}

The classical version of the NNGA introduced in Algorithm~\ref{alg:nnga} requires, for each candidate point that we compute, the distance to all other data points, from which the mean of the $k$ nearest neighbors is set to be the current prototype. The algorithm  associates this prototype with the original candidate points. 
We repeat this step until the prototype moves less than a threshold $\varepsilon_1$ or whenever the algorithm have reached  $j_{\max}$ iterations.

The complexity for the exact nearest neighbors search of a single point is $n\log(n)$. Applied to every data point multiple times, this complexity increases to $n^2 j_{\max} \log(n)$,
preventing its application on Big datasets.


\section{Model proposition }
\subsection{Approximate nearest neighbors search for density gradient ascent}
\label{subsec:lsh}

\begin{figure}[!ht]
\centering
\includegraphics[height=5.5cm,width=0.8\columnwidth]{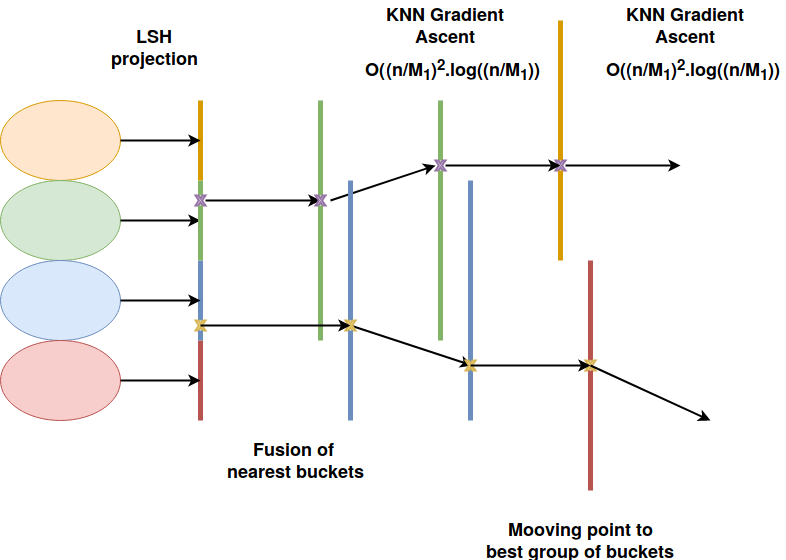} 
\caption{Passage of a prototype through different LSH buckets during the gradient ascent. $n$ and $M_1$ indicate respectively the number of observation and the number of  buckets}
\label{fig:ms-illus2}
\end{figure}

One promising algorithmic complexity reduction approach relies on computing approximate nearest neighbors rather than exact neighbors. 
Among the techniques that can be used, Locality Sensitive Hashing (LSH), introduced in \cite{indyk1998,datar2004}, is a probabilistic method based on a random scalar projection of multivariate data point $\bx$ defined below:
$$ L (\bx; w) = (\boldsymbol{Z}^T \bx + U)/w $$
where $\boldsymbol{Z} \sim N(0, \bI_d)$ is a standard $d$-variate normal random variable and $U \sim \operatorname{Unif}(0, w)$ is a uniform random variable on $[0,w)$, $w >0$.  
The LSH is parametrized by the number of buckets $M_{1}$ in the hash table.  In our context, we propose to set $w=1$, and without loss of generality $L_i \equiv L(\bX_i; 1)$. These
scalar projections are sorted into their order statistics $L_{(1)} < \dots < L_{(n)}$, and their range is divided into $M_{1}$ partition intervals of width $w=(L_{(n)} - L_{(1)})/M_1$ where $I_j = [ L_{(1)} + w(j-1), L_{(1)} + wj]$, $\forall j\in$  $\{j=1, \dots, M_1\}$. The hash value of $\bx$ is the index of the interval in which $L(\bx;1)$ falls 
\begin{equation*}
 H(\bx) = j \indi{L(\bx; 1) \in I_j}, 
\end{equation*}
where $\indi{\cdot}$ is the indicator function. To search for approximate nearest neighbors, the reservoir of potential nearest neighbors is set to the bucket which contains the hash value. This reservoir is enlarged if necessary by concatenating the adjacent buckets.  
The approximate $k$ nearest neighbors of $\bx$ are the $k$ nearest neighbors only drawn from the reduced reservoir $R(\bx)$ defined as below: 
$$ \kann (\bx) = \lbrace \bX_i \in R(\bx) : \lVert \bx - \bX_i \lVert \ \leq \delta_{(k)}(\bx)\rbrace,$$
where $\delta_{(k)}(\bx)$ is the $k$th nearest neighbor distance to $\bx$.
The approximation error in the nearest neighbors to $\bx$ induced by searching in $R(\bx)$ rather than the full dataset is probabilistically controlled \cite{slaney2008}. 
%
%
Our work introduced an improvement to the classical LSH, based on the following observations, as follows:
\begin{itemize}
\item it takes into account  the properties of the local data space more accurately. Rather than looking exclusively in the bucket where the prototype lies, we also look in the neighbor buckets on both sides (i.e. $p=1$).
%
The main advantage of this is to take into account the case where a prototype is at the border of a bucket and so some of its $k$ nearest neighbors mostly likely fall into the neighbor buckets. This is especially important for high values of $k$. It is computationally more expensive but the cost can be controlled control for a fixed bucket size. 
The memory cost is increased  by a factor of $2p + 1$ per partition due to copying the neighboring layers into the active one. 

The LSH method partitions the data space into buckets of approximately $k$ nearest neighbors, which are delimited by parallel hyperplanes. In practice, the LSH controls the number of neighboring buckets to two, except for the edge buckets which have only one neighbor bucket. This is in contrast to cell based buckets, where the number of neighbor buckets increases exponentially with the number of dimensions. Fig.~\ref{fig:LSH-illus} illustrates the LSH buckets of approximate nearest neighbors on 2D and 3D data examples: the orientation of hyperplanes depends on the random projections utilized to construct the buckets.

\begin{figure}[!ht]
\centering
\begin{tabular}{cc} 
\begin{subfigure}[t]{0.40\textwidth}
\includegraphics[width=\columnwidth]{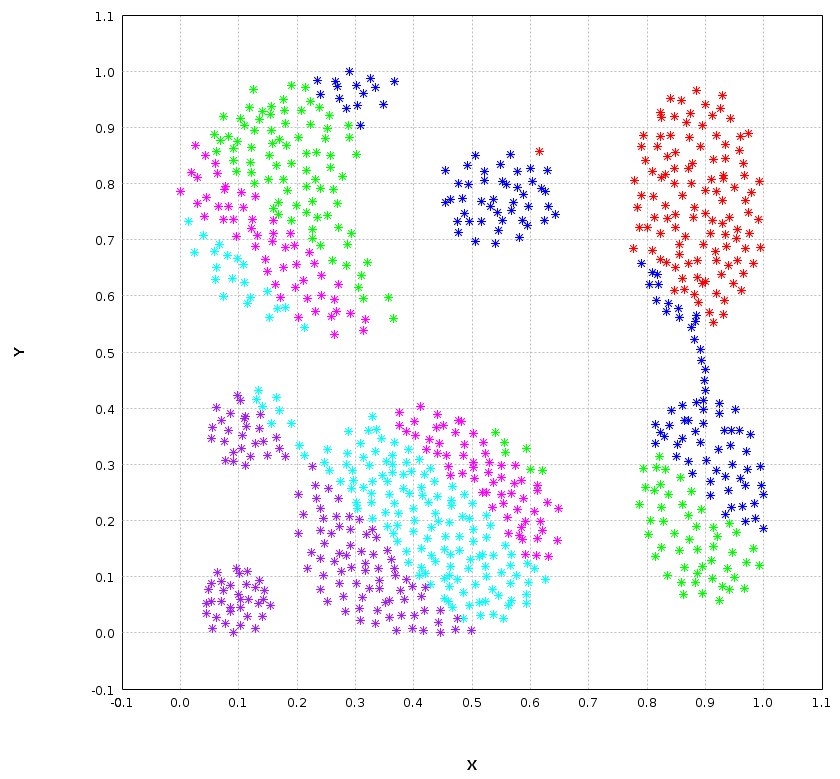}
\caption{Aggregation [$M_1=6$]}
\end{subfigure}
&
\begin{subfigure}[t]{0.37\textwidth}
\includegraphics[width=\columnwidth]{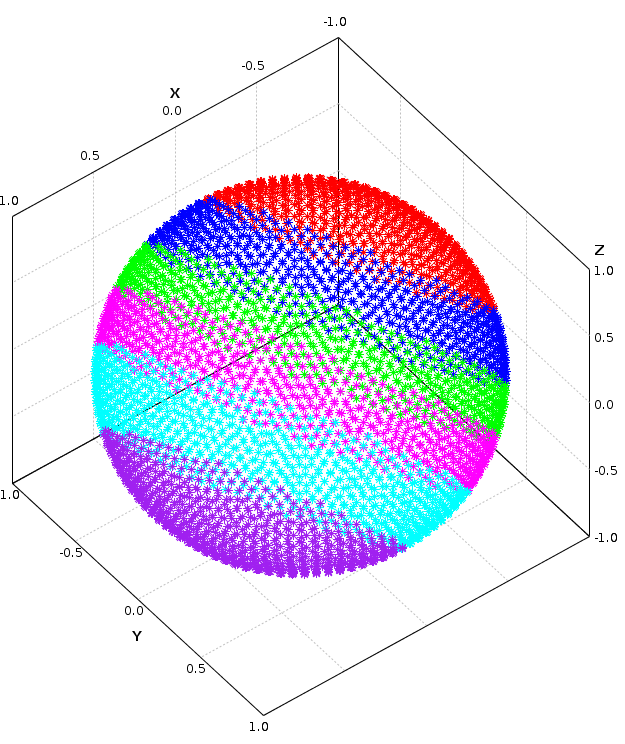}
\caption{GolfBall [$M_1=6$]}
\end{subfigure}
\end{tabular}
\caption{LSH buckets for the Aggregation and GolfBall  datasets. The label $[M_1]$ indicates the LSH with $M_1$ buckets.}
\label{fig:LSH-illus}
\end{figure}

\item One  add the possibility of allowing the prototype to change buckets during its gradient ascent. In this case,  we look for its $k_2$ nearest neighbors in order to place the prototype within the most representative bucket using a majority voting process. Thus a prototype can  pass through multiple buckets before converging to its final position, as illustrated in Fig.~\ref{fig:ms-illus2}.
\end{itemize}

 Algorithm~\ref{alg:nnlshga} describes the NNGA$^{+}$, an approximate nearest neighbor search usin LSH with the hash function $H$. The inputs are the data sample $\bX_1, \dots, \bX_n$, the candidate points $\bx_1, \dots, \bx_m$, and the tuning parameters: the number of nearest neighbors $k_1$ and the number of buckets in the hash table $M_{1}$. In line 1, the hash table is created by applying the LSH to the data values $\bX_1, \dots, \bX_n$. In lines 2--6, for each candidate point $\bx_\ell$, the approximate $k_1$ nearest neighbors $k\textnormal{-}\widetilde{\operatorname{nn}} (\bx_\ell)$ are computed  from within the reservoir $R(\bx_\ell)$.

\begin{algorithm}[!ht]
\begin{algorithmic}[1]
\Statex {\bf Input:} $\lbrace \bX_1, \dots, \bX_n \rbrace, \lbrace \bx_1, \dots, \bx_m \rbrace, k_1, M_1$ 
\Statex {\bf Output:} $\lbrace \kann(\bx_1), \dots, \kann(\bx_m) \rbrace$
\Statex /* Create hash table with $M_1$ buckets */
\LineFor{$i := 1$ to $n$}{$H_i := H(\bX_i)$;}
\Statex /* Search for approx $nn$ in adjacent buckets */
\For{$\ell := 1$ to $m$}
\State $R(\bx_\ell)$ := $\lbrace \bX_i : H_i = H(\bx_\ell), i \in \lbrace 1, \dots, n\rbrace\rbrace$\;
\While{$\operatorname{card}(R(\bx_\ell))< k_1$}
\State $R(\bx_\ell) := R(\bx_\ell) \ \cup$ neighbor bucket;
\EndWhile
\State $\kann(\bx_\ell) := k\textnormal{-nn}$ from $R(\bx_\ell)$ to $\bx_\ell$;
\EndFor
\end{algorithmic}
\caption{NNGA$^{+}$ -- Approximate Nearest Neighbors Gradient Ascent with LSH and adjacent buckets}
\label{alg:nnlshga}
\end{algorithm}

\begin{figure}[!ht]
\centering
\begin{tabular}{@{}c@{}c@{}} 
\begin{subfigure}[t]{0.40\textwidth}
\includegraphics[width=1\columnwidth]{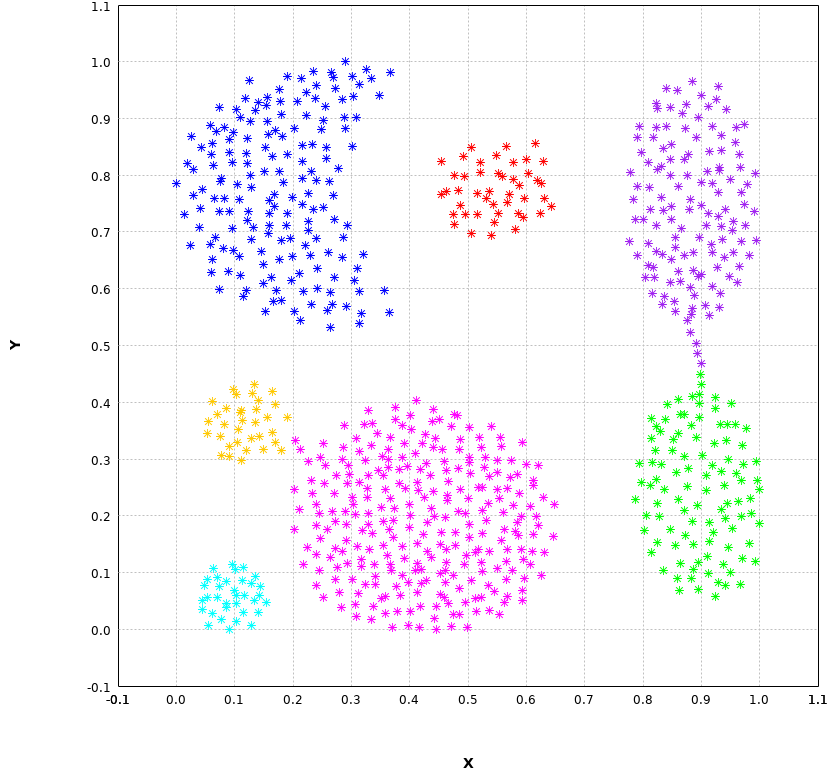}
\caption{Aggregation}
\end{subfigure} &
\begin{subfigure}[t]{0.40\textwidth}
\includegraphics[width=1\columnwidth]{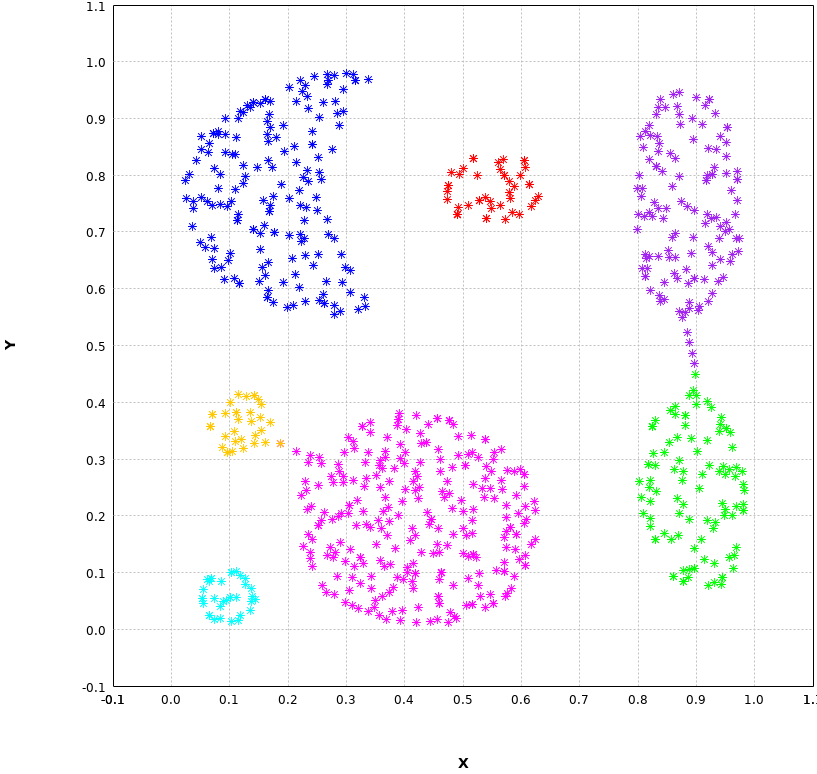}
\caption{Aggregation  $[k_1=5, p=0]$}
\end{subfigure}
\\ \\
\begin{subfigure}[t]{0.40\textwidth}
\includegraphics[width=1\columnwidth]{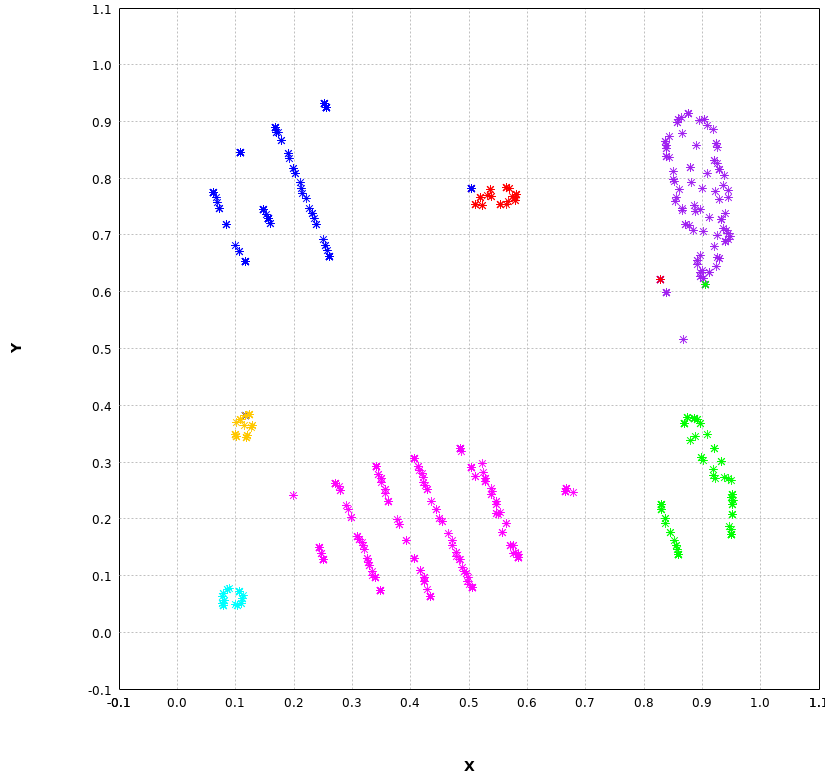}
\caption{Aggregation  $[k_1=20, p=0]$}
\end{subfigure} &
\begin{subfigure}[t]{0.40\textwidth}
\includegraphics[width=1\columnwidth]{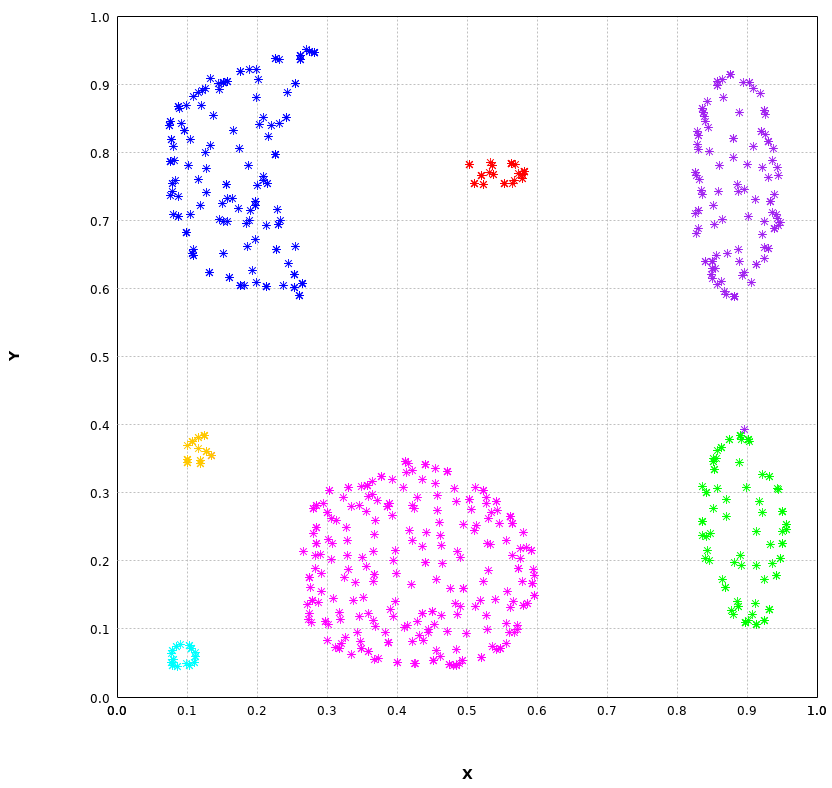}
\caption{Aggregation  $[k_1=20, p=1]$}
\end{subfigure} 
\end{tabular}
\caption{Results of NNGA$^{+}$ for the Aggregation dataset. [$k_1, p]$ indicates NNGA$^{+}$ with $k_1$ nearest neighbors $k_1$ and $p$ neighbor bucket layers.}
\label{fig:practicalGA}
\end{figure}

Fig.\ref{fig:practicalGA} illustrates the effect of including neighbor buckets $(p=1)$ or not ($p=0$) in the NNGA$^{+}$.
Fig.~\ref{fig:practicalGA}a demonstrates the effective of our algorithm on the Aggregation dataset. In Figure~\ref{fig:practicalGA}b  a low value of $k_1$ is used and no neighbor buckets are allowed where the data shrinks slightly. If we increase the value of $k_1$ to $20$ with no neighbor buckets, in Figure~\ref{fig:practicalGA}c,  then we observe that the data are artificially forced to follow the hyperplanes which delimit the different buckets. Figure~\ref{fig:practicalGA}d shows our algorithmic improvement with adding one layer of neighbor buckets where the underlying  structure of data is maintained. 

It is important to note that the use of the LSH to reduce the complexity of kernel Mean Shift clustering was already proposed in \cite{cui2011},  but authors did not quantify the reduction in complexity.  The complexity of our NNGA$^{+}$ is reduced to $O((\frac{n}{M_{1}})^2 \log(\frac{n}{M_{1}}))$ per bucket with $M_1$ buckets, and so the total complexity is $O((\frac{n}{M_{1}})^2 \log(\frac{n}{M_{1}}))$ for all buckets. Because of this segmentation of the original data space into $M_1$ sub-spaces, the complexity is inversely proportional to the number of buckets. The trade-off is that the data points in each bucket have to be sufficiently representative of the local properties of the original space. 
Thus the number of buckets $M_{1}$ is a crucial tuning parameter. Despite this, there are no optimal methods for selecting the number of buckets \cite{har-peled2012}. Consequently, we will examine empirical choices of the number of buckets to study the performance of our proposed method in section \ref{S:2}. 

\subsection{Cluster labeling: $\varepsilon$-proximity}

NNGA$^{+}$ carries the gradient ascent  on the data points until they have converged to their prototypes. The question is then how to label the data point with clusters.
A first solution  is to assign the same cluster to all points sharing the same prototype.
As observed in Figure~\ref{fig:practicalGA}d, even with an a prior good choice of $k_1$, there are (possibly) hundreds of generated prototypes, so assigning a label to each point according to its closest prototype is not effective because of generating too many clusters. 
%
%
Applying the density gradient ascent NNGA$^{+}$
leads to a converged dataset with increased inter-cluster distances and decreased intra-cluster distances as compared to the original dataset. In order to further exploit this property, we propose a new proximity-based approach where points which are under a threshold $\varepsilon$ from each other are considered to belong to the same cluster.
%
%

The aim of $\varepsilon$-proximity algorithm is to gather all points which are under a distance $\varepsilon$ from each other. In order to apply this method, we have to build the similarity matrix which has a $O(n^2)$ time complexity, preventing any Big Data application. For a scalable approach, we apply the LSH on the dataset and generate clusters on these LSH buckets through the local version of $\varepsilon$-proximity clustering.

 Algorithm~\ref{alg:epsilonClustering} illustrates the local version of the algorithm. It consists in exploring the similarity matrix $\mathbf S$ which is defined as a map whose objects IDs are the keys and whose pairs (object IDs, distances) are the values. We initialize the process by taking the first object of $\mathbf S$  and cluster with it every point whose distance is less than $\varepsilon_2$. We then apply this exploration process by iteratively adding he $\varepsilon_2$ nearest neighbors of these added points until this process terminates. During the process we remove the explored points from $\mathbf S$  to avoid repeated calculations. Once the first cluster is generated, we take another object from outside this first cluster from the reduced similarity matrix $\mathbf S$  and repeat the above cluster formation, until all objects are assigned into a cluster label.

The distributed version consists to apply this algorithm in each LSH partitions, we merge with each bucket its right or left neighbor bucket, maintaining the bucket order.  Once this step is completed, we apply a \emph{MapPartitions} procedure where and check if two clusters of two different buckets share $k_3$ pairs of points under $\varepsilon$ (we use $k_3=1$ throughout this paper). Thus these two clusters are considered to form a single cluster. We obtain a dataset which chains common clusters between partitions: all chained clusters are assigned with the same label by generating an undirected graph where each connected subgraph represents a cluster. The search for connected components in a graph is a common problem which can be solved in $O(n)$n $n$ being the number of vertices.

It is important to bound the buckets size because the local version of $\varepsilon$-proximity clustering and the check for cluster fusion between two buckets have quadratic complexity in this size.
Empirically we advise to set the number of buckets in order to have around 500 to 2000 data points in each bucket. 

A notable problem still remains with the choice of the main tuning parameter $\varepsilon$. 
We set it to be the average of distance from each point to their $k$ nearest neighbors. We compute it as an approximate value in using  LSH procedure in order to maintain the scalability property. 


\begin{algorithm}[!ht]
\caption{Local $\varepsilon$-proximity labeling}
\label{alg:epsilonClustering}
\begin{algorithmic}[1]
\Statex {\bf Input:} $\{ \bx_1, \dots, \bx_m \}, {\bf S}, \varepsilon_2$ 
\Statex {\bf Output:} $\{ \tilde{\clust}(\bx_1), \dots, \tilde{\clust}(\bx_m)\}$
\State $\mathtt{needToVisit}$ $\leftarrow$ Set({\bf S}.head)
\State $\mathtt{c_{ID}} \leftarrow 0$
\State $\mathtt{clusters} \leftarrow$ Map.empty[Int, Set[Int]]
\State $\mathtt{clusters}$ += $(\mathtt{c_{ID}}$, $\mathtt{needToVisit}$)
\While{${\bf S}$ has elements}
\Statex /* $p_c$ is the current point of $\mathtt{needToVisit}$ */
\For{{\bf each} $p_c$ {\bf in} $\mathtt{needToVisit}$}
\Statex /* Add all points under $\varepsilon_2$ to needToVisit */
\State $\mathtt{needToVisitUpdated} \leftarrow \{p \in {\bf S}$, $\mathrm{dist}(p, \mathrm{p_c}) \leq \varepsilon_2\}$
\Statex /* Remove explored point from the similarity matrix */
\State ${\bf S}$ -= $p_c$
\Statex /* Update points to explore on next iterations */
\State $\mathtt{needToVisit} \leftarrow \mathtt{needToVisitUpdated}$
\If{ $\mathtt{neetToVisit}$ is empty }
\State $\mathtt{c_{ID}}$ += 1
\State $\mathtt{needToVisit} \leftarrow$ Set({\bf S}.head)
\Statex /* Create a new entry in the clusters map */
\State $\mathtt{clusters}$ += ($\mathtt{c_{ID}}$, $\mathtt{needToVisit}$)
\Else
\Statex /* Add new points to cluster $c_{ID}$ */
\State $\mathtt{clusters(c_{ID})}$ += $\mathtt{needToVisit}$
\EndIf
\EndFor
\EndWhile
\end{algorithmic}
\end{algorithm}

\section{Numerical experiments}
\label{S:2}

Our experiments are carried out on the Grid'5000 testbed which is the french national testbed for computer science research. It allows the deployment of a user-specified operating system within the Grid'5000 hardware. We use a dedicated Spark Linux image
optimized for Grid'5000 where Apache Spark is deployed on top of Spark in Standalone mode.  
Apache Spark is a fast general purpose distributed computing system based on a master-slaves architecture. Only the deployment of the image is automatized. We manually reserve the nodes and provide the Spark cluster with our code
to execute the different experiments on a $2 \times 8$ core Intel Xeon E5-2630v3 CPUs and 128 Gb RAM set up. We repeat each experiment ten times for robustness. 


A key concept in Spark is the resilient distributed dataset (RDD) which is a read-only collection of objects partitioned across a group of machines which can be rebuilt if necessary from the hierarchy of previous RDD operations. 
 Most of the $Map$ and $Reduce$ operations will be performed on RDDs even if other pure Scala $Map$ and $Reduce$ operations are executed inside each Spark partition. 
We implement our algorithm in Scala because it is the Spark's  native language and thus allows for good performance.  
Most  tuning parameters of our algorithm have an impact on both the execution time and the cluster labeling quality. We focus our study on the dataset size $n$, the maximum number 
of Gradient Ascent
iterations $j_{\max}$, the number of LSH buckets $M_1$, the threshold 
$\varepsilon_1$, and the number of nearest neighbors $k_1$. 
\subsection{Datasets and tuning parameters}
We use a range from  2  to high dimensional datasets with different sizes, as summarized in Table~\ref{tab:Datasets} \cite{Ultsch2005, ClusteringDatasets2015}.
To ensure the comparability of the results across these different datasets, all algorithms are carried out on the normalized version of the datasets:
$x_i = (x_i - x_i^{\min})/(x_i^{\max}-x_i^{\min}) $
where $x_i$ is the $i$th component of  $\bx$, and $x_i^{\min}, x_i^{\max}$ are respectively the $i$th marginal minimum and maximum values.  Table \ref{tab:clustAlgoParam} shows the combination of tuning parameters used in the comparision between the gradient ascent (NNGA$^{+}$), $k$-means \cite{DBLP:journals/corr/abs-1203-6402}, DBScan \cite{dbscanDistributed} and $\varepsilon$-proximity. 
%
\begin{table}[htpb]
\centering
\begin{tabular}{|l|r|r|r|}
\hline
Dataset & $n$ & $d$ & $N$ \\\hline
R15 & 600 & 2 & 15         \\
Aggregation & 788 & 2 & 7  \\
Sizes5 & 1000 & 2 & 4     \\ 
EngyTime & 4096 & 2 & 2   \\ 
Banana & 4811 & 2 & 2 	\\
S3 & 5000 & 2 & 15  \\
Disk6000 & 6000 & 2 & 2 \\
DS1 & 9153 & 2 & 14 \\
Hepta & 212 & 3 & 7 \\
Hyperplane & 100000 & 10 & 5\\
CovType10 & 581012 & 10 & 7 \\
image\footnote{http://www.eecs.berkeley.edu/Research/Projects/CS/vision/bsds} & 154401 & 5 & -- \\
Own image & 5000000 & 5 & -- \\

ScalabiltyDS & 140000000 & 10 & -- \\

\hline
\end{tabular}
\caption{Experimental datasets. $n$ is the dataset size, $d$ is the data dimension, $N$ is the number of clusters. }
\label{tab:Datasets}
\end{table}

\begin{table}[htbp]
\centering
\scalebox{0.6}{
\begin{tabular}{|l|l|c|c|c|}
\hline
dataset & NNGA$^{+}$ & $k$-means & DBScan & $\varepsilon$-proximity \\ \hline
Aggregation & without & $k=7$ & $\varepsilon=0.05,$ & $\varepsilon_{knn}=10,$ \\
& & & $minPts=8$ & $M_1=8$ \\
& with $k_1=40$ & & & $\varepsilon_{knn}=30$,\\
& & & & $M_1=8$ \\\hline
Banana & without & $k=2$ & $\varepsilon=0.02,$ & $\varepsilon_{knn}=0.1,$ \\
& & & $minPts=3$ & $M_1=8$ \\
& with $k_1=40$ & & & \\\hline
Disk6000 & without & $k=7$ & $\varepsilon=0.02,$ & $\varepsilon_{knn}=0.021,$ \\
& & & $minPts=4$ & $M_1=8$ \\
& with $k_1=40$ & & & \\\hline
DS1 & without & $k=14$ & $\varepsilon=0.03,$ & $\varepsilon_{knn}=30,$ \\
& & & $minPts=25$ & $M_1=8$ \\
& with $k_1=40$ & & & \\\hline
EngyTime & without & $k=2$ & $\varepsilon=0.1,$ & $\varepsilon_{knn}=50,$ \\
& & & $minPts=100$ & $M_1=8$ \\
& with $k_1=40$ & & & \\\hline
Hepta & without & $k=7$ & $\varepsilon=0.1,$ & $\varepsilon_{knn}=10,$ \\
& & & $minPts=10$ & $M_1=4$ \\
& with $k_1=40$ & & & $\varepsilon_{knn}=20$,\\
& & & & $M_1=4$ \\\hline
R15 & without & $k=15$ & $\varepsilon=0.05,$ & $\varepsilon_{knn}=5,$ \\
& & & $minPts=25$ & $M_1=8$ \\
& with $k_1=40$ & & & \\\hline
S3 & without & $k=15$ & $\varepsilon=0.05,$ & $\varepsilon_{knn}=15,$ \\
& & & $minPts=50$ & $M_1=8$ \\
& with $k_1=40$ & & & $\varepsilon_{knn}=15$,\\
& & & & $M_1=8$ \\\hline
Sizes5 & without & $k=4$ & $\varepsilon=0.08,$ & $\varepsilon_{knn}=10,$ \\
& & & $minPts=8$ & $M_1=8$ \\
& with $k_1=40$ & & & $\varepsilon_{knn}=35$,\\
& & & & $M_1=8$ \\\hline
Unbalanced & without & $k=8$ & $\varepsilon=0.05,$ & $\varepsilon_{knn}=50,$ \\
& & & $minPts=20$ & $M_1=8$ \\
& with $k_1=40$ & & $\varepsilon=0.1,$ & $\varepsilon_{knn}=80$,\\
& & & $minPts=20$ & $M_1=8$ \\\hline
Hyperplane & without & $k=5$ & $\varepsilon=0.05,$ & $\varepsilon_{knn}=5,$ \\
& & & $minPts=8$ & $M_1=100$ \\
& with $k_1=40$ & & & \\\hline
CovType10 & without & $k=5$ & $\varepsilon=0.05,$ & $\varepsilon_{knn}=20,$ \\
& & & $minPts=8$ & $M_1=500$ \\ \hline
\end{tabular}
}
\caption{Clustering algorithms parameters. $\varepsilon=v_1$ stands for manual setting of $\varepsilon$ and $\varepsilon_{knn}=v_2$ defines the number of the $k$ nearest neighbor for the $\varepsilon$ approximation. $M_1=$ represents the number of buckets during the LSH phase.}
\label{tab:clustAlgoParam}
\end{table}
\subsection{Evaluation of clustering quality}
\subsubsection{Quantitative evaluation}
NNGA$^{+}$ is compared to the following cluster labeling techniques:  $k$-means \cite{DBLP:journals/corr/abs-1203-6402}, 
DBScan \cite{dbscanDistributed},
and $\varepsilon$-proximity. To evaluate the quality of the clustering, we use both the 
Normalized Mutual Information (NMI) \cite{nmi05} and the RAND index \cite{conf/icann/SantosE09}.
The value of each measure lies between 0 and 1. A higher value indicates better
clustering results.

One notable observation of the impact of the NNGA$^{+}$ pre-processing step on clustering labeling concerns the datasets with Gaussian (i.e. ellipsoidal) clusters. As the results for the Hepta, R15, S3, and Sizes5 datasets as Table~\ref{tab:idxComp1} and \ref{tab:idxComp2} show, applying NNGA$^{+}$ results in uniformly better clustering quality than without the NNGA$^{+}$. NNGA$^{+}$ pre-processing in DBScan and $\varepsilon$-proximity clustering also leads to better (or at least as good) clustering accuracy than without NNGA$^{+}$ except for Disk6000 which has nested clusters. In this case, a high $k_1$ value leads to the creation of links between clusters which are well separated. Finally we observe for the high dimensional datasets, Hyperplan and CovType10, $\varepsilon$-proximity with NNGA$^{+}$ pre-processing performs as well as the classical $k$-means, and it out-performs the DBScan which is unable to effectively cluster such huge datasets. 


\begin{table}[htbp]
\centering
\scalebox{0.65}{
\begin{tabular}{|l|l|c|c|c|}
\hline
Data & NNGA$^{+}$  & $k$-means & DBScan & $\varepsilon$-proximity \\ \hline
Aggregation & without & $0.83 \pm 0.022$ & $0.98 \pm 0.00$ &$0.89 \pm 0.00$ \\
& with [$k_1=50$]  & $\mathbf{0.87 \pm 0.06}$ & $\mathbf{0.99 \pm 0.00}$ & $\mathbf{0.97 \pm 0.02}$ \\\hline
Banana & without & $0.31 \pm 0.00$ & $\mathbf{1.00 \pm 0.00}$ & $\mathbf{1.00 \pm 0.00}$ \\
& with [$k=40$] & $\mathbf{0.32 \pm 0.01}$ & $\mathbf{1.00 \pm 0.00}$ & $\mathbf{1.00 \pm 0.00}$ \\\hline
Disk6000 & without & $0.00 \pm 0.00$ & $\mathbf{1.00 \pm 0.00}$ & $\mathbf{1.00 \pm 0.00}$ \\
& with [$k=100$] & $\mathbf{0.02 \pm 0.00}$ & $0.25 \pm 0.00$ & $0.32 \pm 0.00$ \\\hline
DS1 & without & $0.75 \pm 0.01$ & $\mathbf{0.94 \pm 0.00}$ & $\mathbf{0.97 \pm 0.00}$ \\
& with [$k_1=50$] &$\mathbf{0.77 \pm 0.03}$ & $\mathbf{0.94 \pm 0.00}$ & $0.95 \pm 0.00$ \\\hline
EngyTime & without & $\mathbf{0.98 \pm 0.00}$ & $0.75 \pm 0.00$ & $0.01 \pm 0.00$\\
& with [$k_1=200$] & $0.97 \pm 0.00$ & $\mathbf{0.97 \pm 0.00}$& $\mathbf{0.85 \pm 0.08}$ \\\hline
Hepta & without  & $0.98 \pm 0.03$ & $0.83 \pm 0.00$ & $\mathbf{0.99 \pm 0.02}$ \\
& with [$k_1=20$] & $\mathbf{0.99 \pm 0.02}$ & $\mathbf{1.00 \pm 0.00}$ & $0.97 \pm 0.04$ \\\hline
R15 & without &  $0.96 \pm 0.02$ & $0.99 \pm 0.00$ & $\mathbf{0.93 \pm 0.01}$ \\
& with [$k_1=20$] & $\mathbf{0.99 \pm 0.00}$ & $\mathbf{0.99 \pm 0.00}$ & $0.91 \pm 0.02$\\\hline
S3 & without & $0.78 \pm 0.01$ & $0.42 \pm 0.00$& $0.09 \pm 0.00$ \\
& with [$k_1=40$] & $\mathbf{0.79 \pm 0.00}$ & $\mathbf{0.75 \pm 0.00}$ & $\mathbf{0.74 \pm 0.00}$ \\\hline
Sizes5 & without & $0.81 \pm 0.12$ & $0.80 \pm 0.00$ & $0.60 \pm 0.12$ \\
& with [$k_1=20$] & $\mathbf{0.89 \pm 0.08}$ & $\mathbf{0.91 \pm 0.00}$ & $\mathbf{0.89 \pm 0.01}$ \\\hline
Unbalanced & without & $\mathbf{0.94 \pm 0.05}$ & $\mathbf{0.99 \pm 0.00}$ & $0.97 \pm 0.01$\\
& with [$k_1=40$] & $0.92 \pm 0.05$ & $\mathbf{0.99 \pm 0.00}$ & $\mathbf{0.98 \pm 0.01}$ \\\hline
Hyperplane & without & $\mathbf{0.01 \pm 0.00}$ & One cluster  & $0.02 \pm 0.00$\\
& with [$k_1=50$] & $0.00 \pm 0.00$ & found & $\mathbf{0.04 \pm 0.00}$ \\\hline
CovType10 & without & $0.07 \pm 0.006$ & dataset is & $0.03 \pm 0.00$\\
& with [$k_1=50$] & $\mathbf{0.07 \pm 0.02}$ & too massive & $\mathbf{0.09 \pm 0.01}$\\\hline
\end{tabular}
}
\caption{NMI clustering quality indices for the cluster labeling on the experimental datasets with and without prior application of NNGA$^{+}$. The bold entries indicate the optimal dataset, and $\pm$ entries are the standard deviations over 10 trials. 
}
\label{tab:idxComp1}
\end{table}

\begin{table}[!ht]
\centering
\scalebox{0.65}{
\begin{tabular}{|l|l|c|c|c|}
\hline
Data & NNGA$^{+}$ & $k$-means & DBScan & $\varepsilon$-proximity \\ \hline
Aggregation & without  & $0.91 \pm 0.01$ & $0.99 \pm 0.00$ & $0.93 \pm 0.00$ \\
& with [$k_1=50$] & $\mathbf{0.93 \pm 0.04}$ & $\mathbf{1.00 \pm 0.00}$ & $\mathbf{0.98 \pm 0.02}$ \\\hline
Banana & without  & $0.70 \pm 0.00$ & $\mathbf{1.00 \pm 0.00}$ & $\mathbf{1.00 \pm 0.00}$ \\
& with [$k=40$]  & $\mathbf{0.70 \pm 0.01}$ & $1.00 \pm 0.00$ & $1.00 \pm 0.00$ \\\hline
Disk6000 & without & $\mathbf{0.50 \pm 0.00}$ & $\mathbf{1.00 \pm 0.00}$ & $\mathbf{1.00 \pm 0.00}$ \\
& with [$k=100$] & $0.50 \pm 0.00$ & $0.33 \pm 0.00$ & $0.34 \pm 0.00$ \\\hline
DS1 & without & $0.86 \pm 0.00$ & $0.98 \pm 0.00$ & $0.98 \pm 0.00$ \\
& with [$k_1=50$] & $\mathbf{0.87 \pm 0.01}$ & $\mathbf{0.98 \pm 0.00}$ & $\mathbf{1.00 \pm 0.00}$ \\\hline
EngyTime & without & $\mathbf{1.00 \pm 0.0}$ & $0.90 \pm 0.00$ & $0.50 \pm 0.00$\\
& with [$k_1=200$] & $0.99 \pm 0.00$ & $\mathbf{0.99 \pm 0.00}$ & $\mathbf{0.93 \pm 0.05}$ \\\hline
Hepta & without & $0.99 \pm 0.02$ & $0.94 \pm 0.00$ & $\mathbf{0.99 \pm 0.02}$ \\
& with [$k_1=20$] & $\mathbf{0.99 \pm 0.02}$ & $\mathbf{1.00 \pm 0.00}$ & $0.98 \pm 0.04$ \\\hline
R15 & without & $0.99 \pm 0.01$ & $1.00 \pm 0.00$ & $\mathbf{0.99 \pm 0.00}$ \\
& with [$k_1=20$] & $\mathbf{1.00 \pm 0.00}$ & $\mathbf{1.00 \pm 0.00}$ & $0.98 \pm 0.00$ \\\hline
S3 & without & $0.96 \pm 0.00$ & $0.52 \pm 0.00$ & $0.12 \pm 0.00$ \\
& with [$k_1=40$] & $\mathbf{0.96 \pm 0.00}$ & $\mathbf{0.95 \pm 0.00}$ & $\mathbf{0.96 \pm 0.00}$ \\\hline
Sizes5 & without & $0.88 \pm 0.13$ & $0.97 \pm 0.00$ & $0.83 \pm 0.07$ \\
& with [$k_1=20$] & $\mathbf{0.95 \pm 0.09}$ & $\mathbf{0.98 \pm 0.00}$ & $\mathbf{0.97 \pm 0.00}$ \\\hline
Unbalanced & without & $\mathbf{0.97 \pm 0.03}$ & $\mathbf{1.00 \pm 0.00}$ & $\mathbf{1.00 \pm 0.00}$\\
& with [$k_1=40$] & $0.96 \pm 0.03$ & $\mathbf{1.00 \pm 0.00}$ & $\mathbf{1.00 \pm 0.00}$ \\\hline
Hyperplane & without & $\mathbf{0.62 \pm 0.00}$ & One cluster& $0.30\pm 1.00$ \\
& with [$k_1=50$] & $0.62 \pm 0.00$ & found & $\mathbf{0.31 \pm 0.00}$ \\\hline
CovType10 & without  & $\mathbf{0.59 \pm 0.00}$ & dataset is & $0.38 \pm 0.00$\\
& with [$k_1=50$] & $0.58 \pm 0.01$ & too massive & $\mathbf{0.56 \pm 0.04}$ \\\hline
\end{tabular}
}
\caption{RAND clustering quality indices for the cluster labeling on the experimental datasets with and without prior application of NNGA$^{+}$. The bold entries indicate the optimal dataset, and $\pm$ entries indicate  the standard deviations over 10 trials. 
}
\label{tab:idxComp2}
\end{table}

\begin{figure}[!ht]
\centering
\setlength{\tabcolsep}{2pt}
\begin{tabular}{@{}cc@{}} 
\begin{subfigure}[t]{0.36\textwidth}
\includegraphics[width=1\columnwidth]{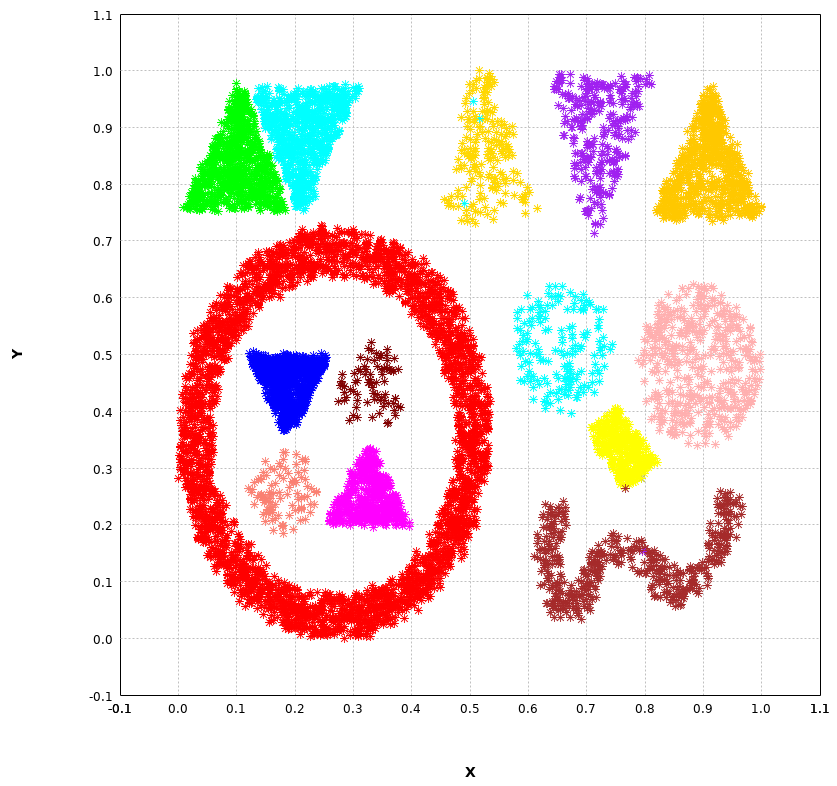} 
\caption{DS1}
\end{subfigure} &
\begin{subfigure}[t]{0.36\textwidth}
\includegraphics[width=1\columnwidth]{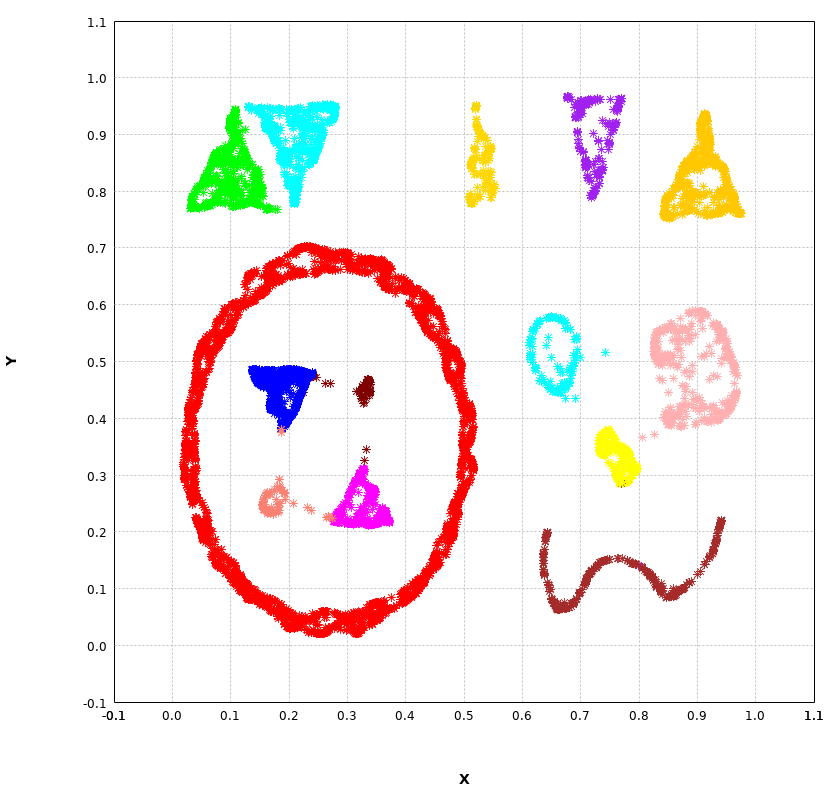}
\caption{NNGA$^{+}$ $[k_1=50]$}
\end{subfigure}
\\
\begin{subfigure}[t]{0.36\textwidth}
\includegraphics[width=1\columnwidth]{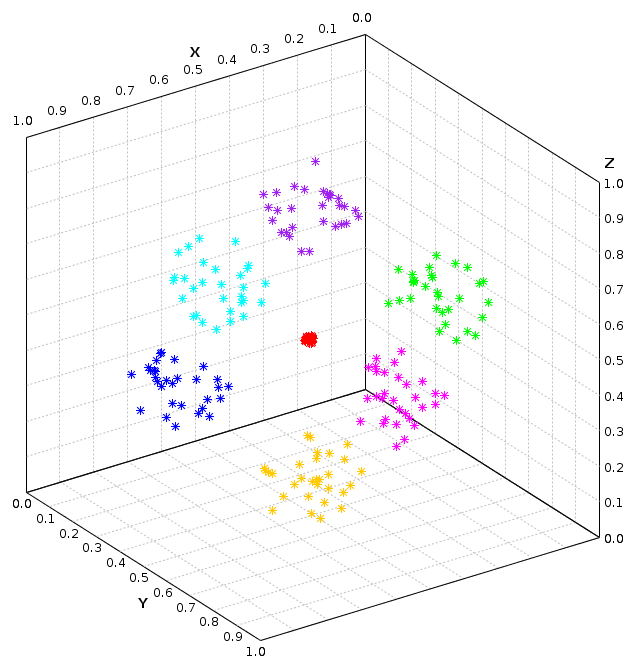}
\caption{Hepta}
\label{hepta_o}
\end{subfigure} &
\begin{subfigure}[t]{0.36\textwidth}
\includegraphics[width=1\columnwidth]{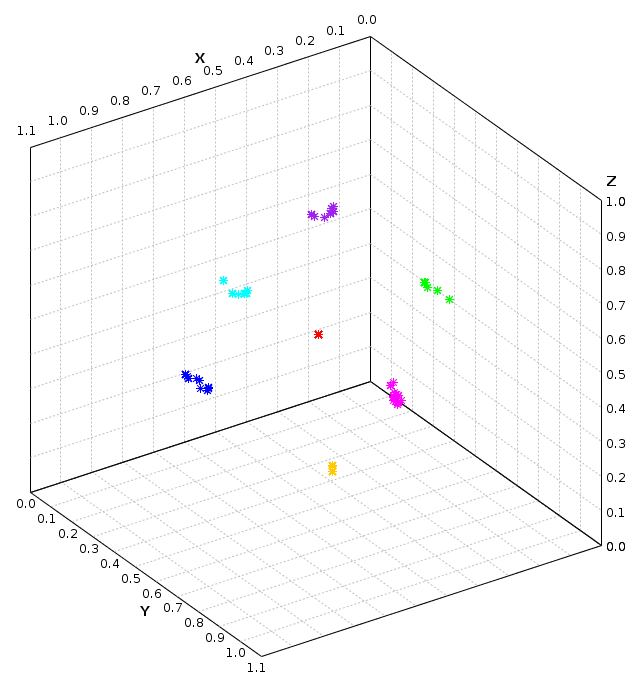}
\caption{NNGA$^{+}$ $[k_1=20]$}
\label{hepta_c20}
\end{subfigure}
\\
\begin{subfigure}[t]{0.36\textwidth}
\includegraphics[width=1\columnwidth]{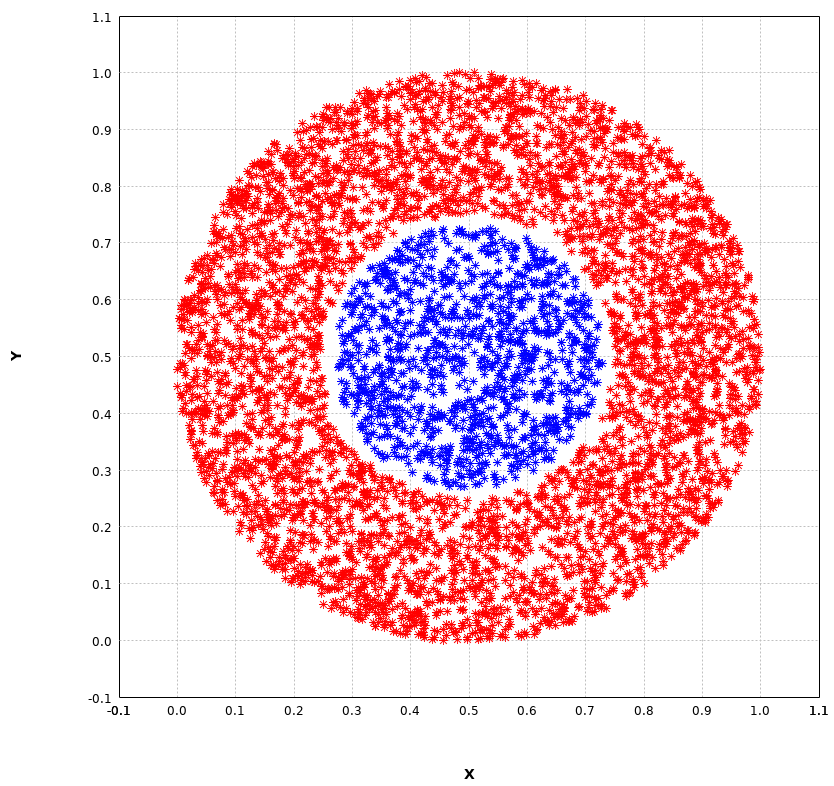}
\caption{Disk6000}
\end{subfigure}
&
\begin{subfigure}[t]{0.36\textwidth}
\includegraphics[width=1\columnwidth]{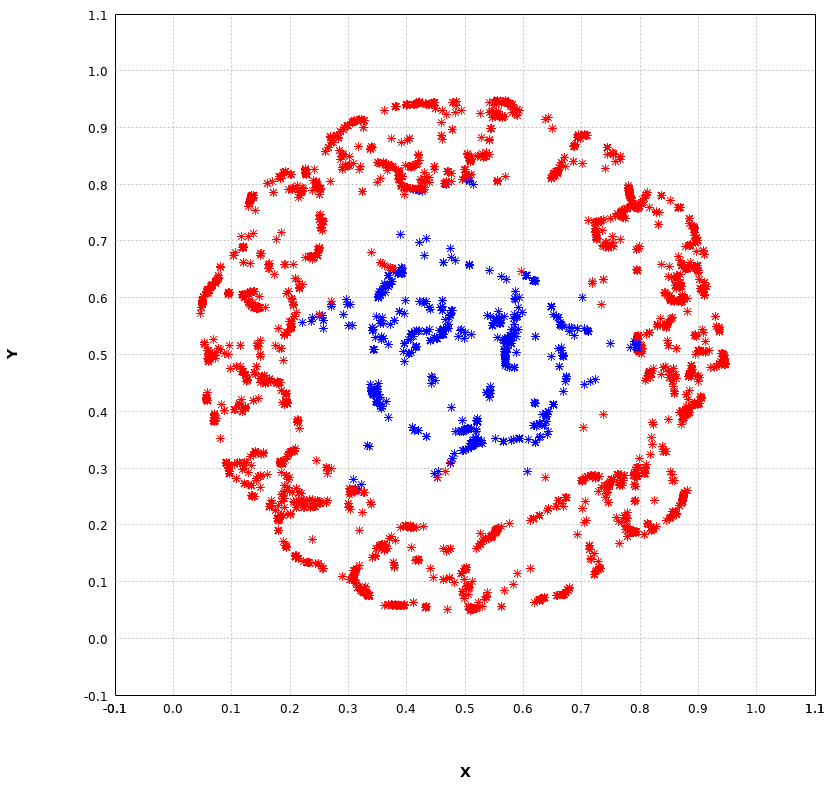}
\caption{NNGA$^{+}$ [$k_1=100$]}
\end{subfigure} 
\end{tabular}
\caption{Gradient ascent NNGA$^{+}$ on the DS1, 
 Hepta and Disk6000 datasets. The label `NNGA$^{+}$ $[k_1]$' indicates the application of NNGA$^{+}$ with $k_1$ nearest neighbors.} 
\label{fig:GAdatasets}
\end{figure}

As shown in Figure.~\ref{fig:GAdatasets}, our version of the nearest neighbors gradient ascent NNGA$^{+}$ results in shrinking the data points toward their local modes. The notation `with NNGA$^{+}$ $[k_1]$' indicates that we carried out NNGA$^{+}$ $k_1$ nearest neighbors in the Mean Shift gradient ascent. DS1 with NNGA$^{+}$ $[k_1=50]$ presents a more compact version than its original version, maintaining the underlying data structures whilst increasing empty space between clusters. Likewise for the Hepta dataset.
One notable difference arises for the uniform clusters in Disk6000. In the latter, we observe formation of empty space next to high density areas, whereas in the former, the data points gather towards the central modes.

An application of NNGA$^{+}$ could result in one of two major cases depending on the nature of clusters. If the points are homogeneously distributed over a cluster, a higher value of $k_1$ results in more empty space surrounding the high density regions. On the other hand, if the points are Gaussian distributed, they will converge efficiently towards the high density region, reducing the noise whilst increasing the inter-cluster distances. 

\begin{figure}[!ht]
\centering
\setlength{\tabcolsep}{2pt}
\begin{tabular}{@{}cc@{}} 
\begin{subfigure}[t]{0.35\textwidth}
\includegraphics[width=1\columnwidth]{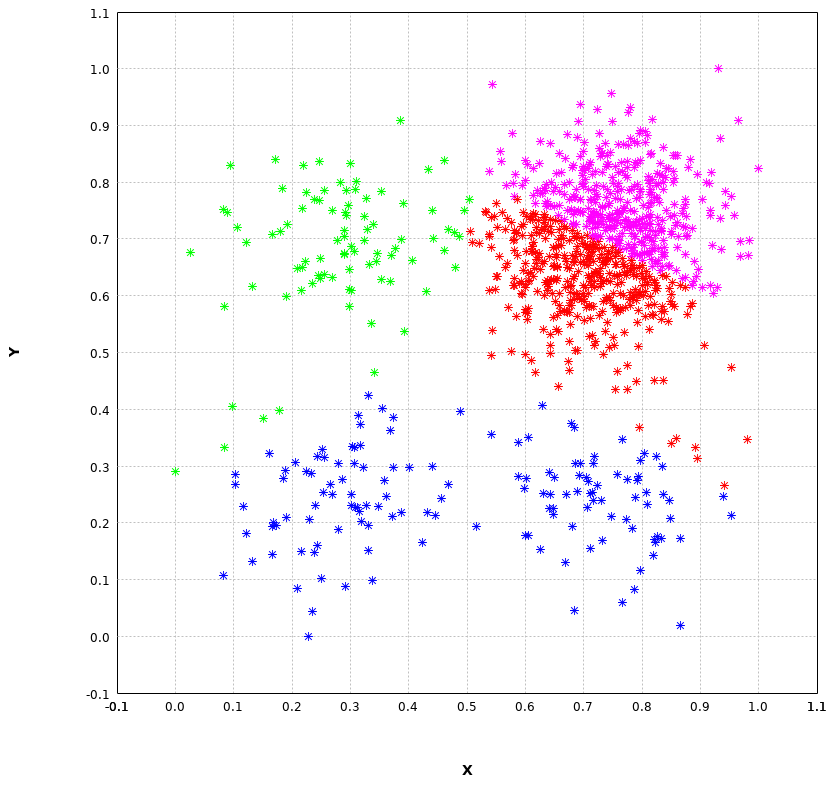}
\caption{$k$-means $[k=4]$}
\label{sizes5_km4}
\end{subfigure}
&
\begin{subfigure}[t]{0.35\textwidth}
\includegraphics[width=1\columnwidth]{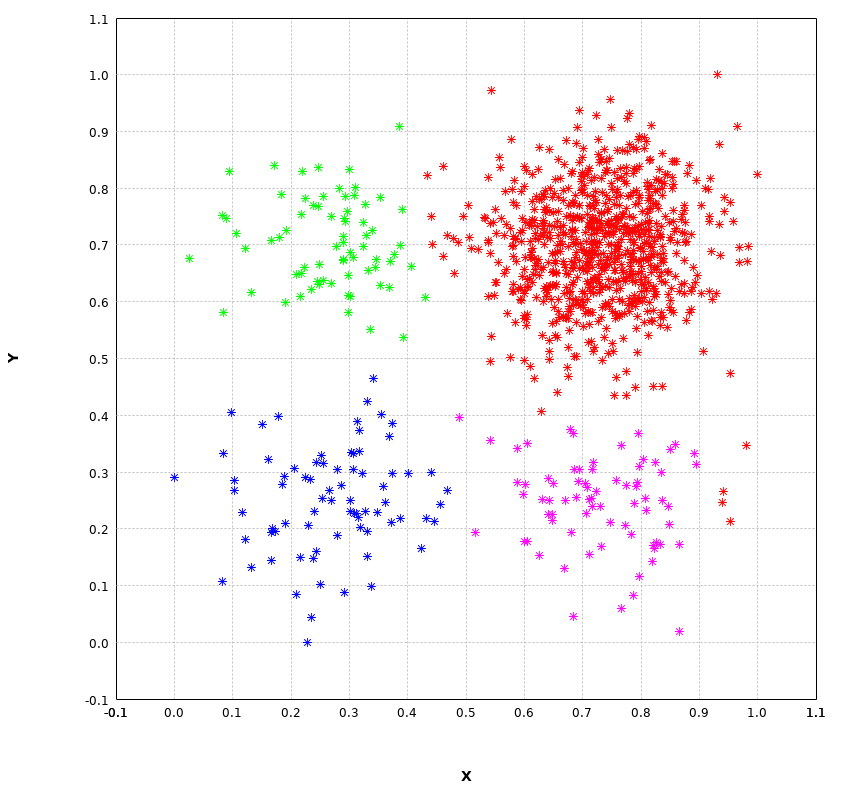}
\caption{NNGA$^{+}$+$k$-means $[k_1=20, k=4]$}
\label{sizes5_c20_o_km4}
\end{subfigure} 
\end{tabular}
\caption{$k$-means cluster labeling on Sizes5 datasets with and without prior application of NNGA$^{+}$. The label `$k$-means $[k]$' indicates $k$-means cluster labeling with $k$ clusters without NNGA$^{+}$, and `NNGA$^{+}$+$k$-means $[k_1, k]$' indicates  NNGA$^{+}$ with $k_1$ nearest neighbors followed by $k$-means cluster labeling with $k$ clusters.}
\label{fig:kmeans-multipleDS}
\end{figure}

Figure~\ref{fig:kmeans-multipleDS} presents some results obtained by applying $k$-means on the original data points and converged data points obtained after applying the NNGA$^{+}$ algorithm. Figure ~\ref{sizes5_km4} shows an application of $k$-means on Sizes5 dataset with $k=4$ clusters without NNGA$^{+}$.  Even if the number of clusters in the $k$-means is the correct number, these four clusters do not match so closely the original clusters (NMI=0.81, RAND=0.88). Figure~\ref{sizes5_c20_o_km4} shows the $k$-means cluster labeling results applied on  NNGA$^{+}$ output with $k_1=20$. We observe visually that these clusters    
are  more similar to the original clusters (NMI=0.89, RAND=0.95). 

\begin{figure}[htbp]
\centering
\setlength{\tabcolsep}{2pt}
\begin{tabular}{@{}cc@{}} 
\begin{subfigure}[t]{0.35\textwidth}
\includegraphics[width=1\columnwidth]{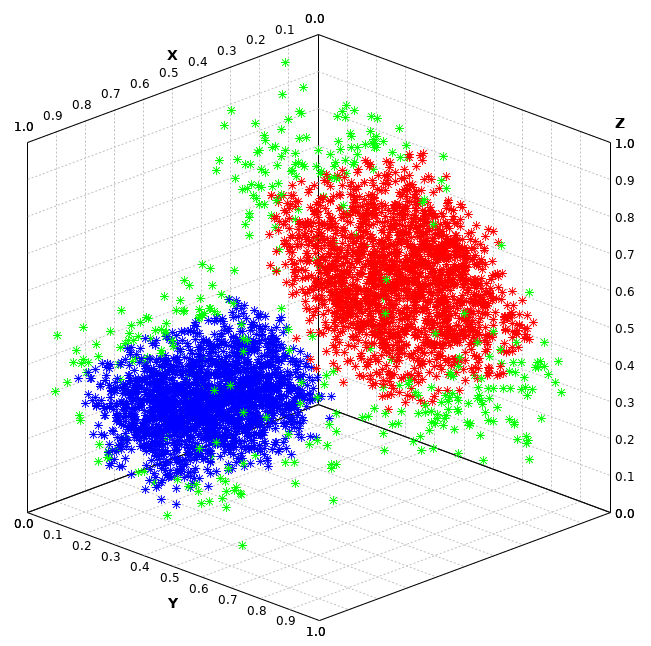}
\caption{DBScan on EngyTime }
\label{dbscanOnEngytTime} 
\end{subfigure}
&
\begin{subfigure}[t]{0.35\textwidth}
\includegraphics[width=1\columnwidth]{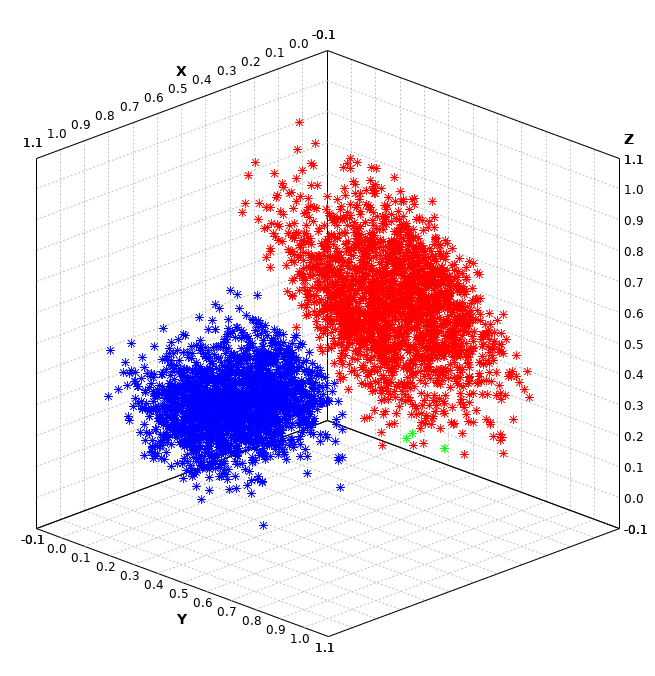}
\caption{NNGA$^{+}$+DBScan $[k_1=200]$}
\label{dbscanOnEngytTimek1_c200o} 
\end{subfigure}
\\
\begin{subfigure}[t]{0.35\textwidth}
\includegraphics[width=1\columnwidth]{pictures/disk6k-original.png}
\caption{DBScan on Disk6000}
\end{subfigure} & 
\begin{subfigure}[t]{0.35\textwidth}
\includegraphics[width=1\columnwidth]{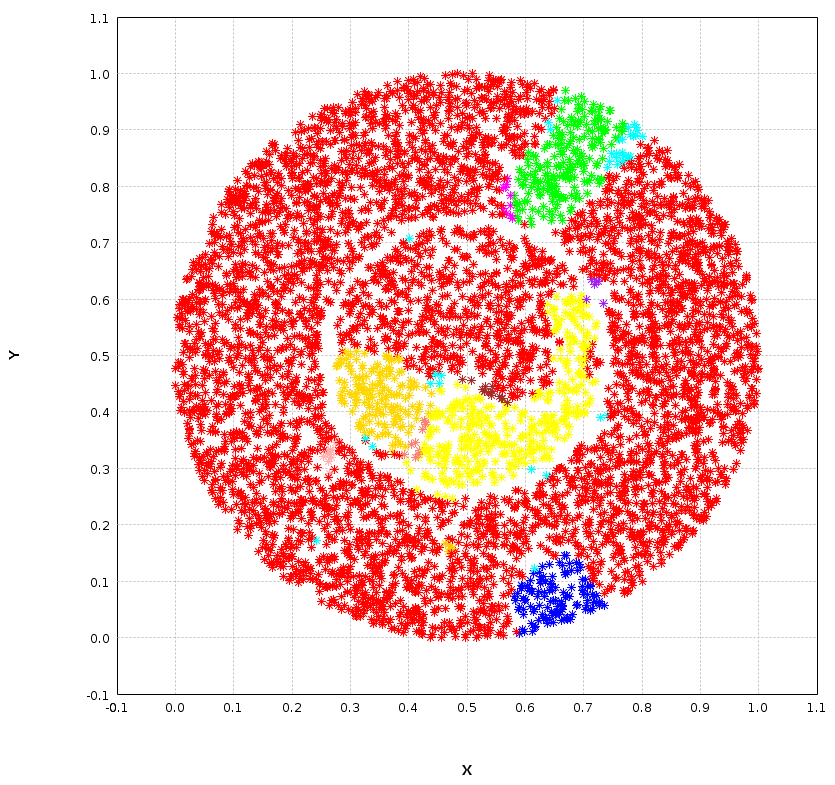}
\caption{NNGA$^{+}$+DBScan $[k_1=100$]}
\end{subfigure}
\end{tabular}
\caption{DBScan labeling on the EngyTime and Disk6000 datasets with and without prior application of NNGA$^{+}$. The label `DBScan' indicates DBScan cluster labeling without NNGA$^{+}$, and `NNGA$^{+}$+DBScan $[k_1]$' indicates  NNGA$^{+}$ with $k_1$ nearest neighbors followed by DBScan cluster labeling.}
\label{fig:dbscanOnDS}
\end{figure}

DBScan cluster labeling collates points more efficiently after NNGA$^{+}$ is applied than without NNGA$^{+}$, as shown in Figure~\ref{fig:dbscanOnDS}. NNGA$^{+}$ is an efficient way to attach noisy points to their closest cluster. 
Furthermore, NNGA$^{+}$ facilitates more robust choices for the DBScan parameters, since it increases the local density which improves the detection of smaller clusters. 

\begin{figure}[htbp]
\centering
\setlength{\tabcolsep}{2pt}
\begin{tabular}{@{}cc@{}} 
\begin{subfigure}[t]{0.35\textwidth}
\includegraphics[width=1\columnwidth]{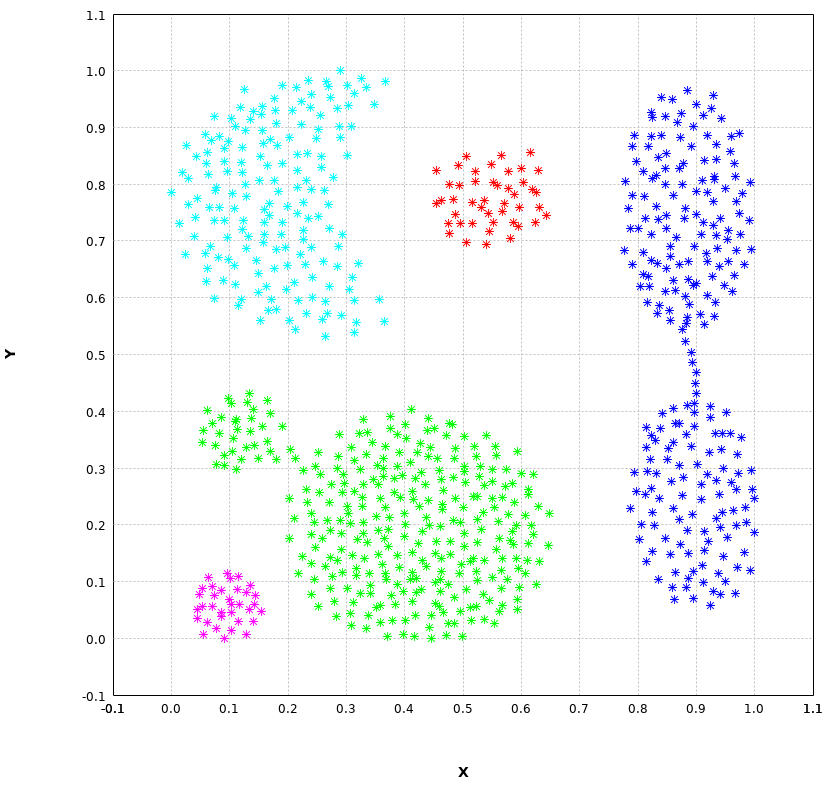}
\caption{$\varepsilon$-proximity}
\end{subfigure} &
\begin{subfigure}[t]{0.35\textwidth}
\includegraphics[width=1\columnwidth]{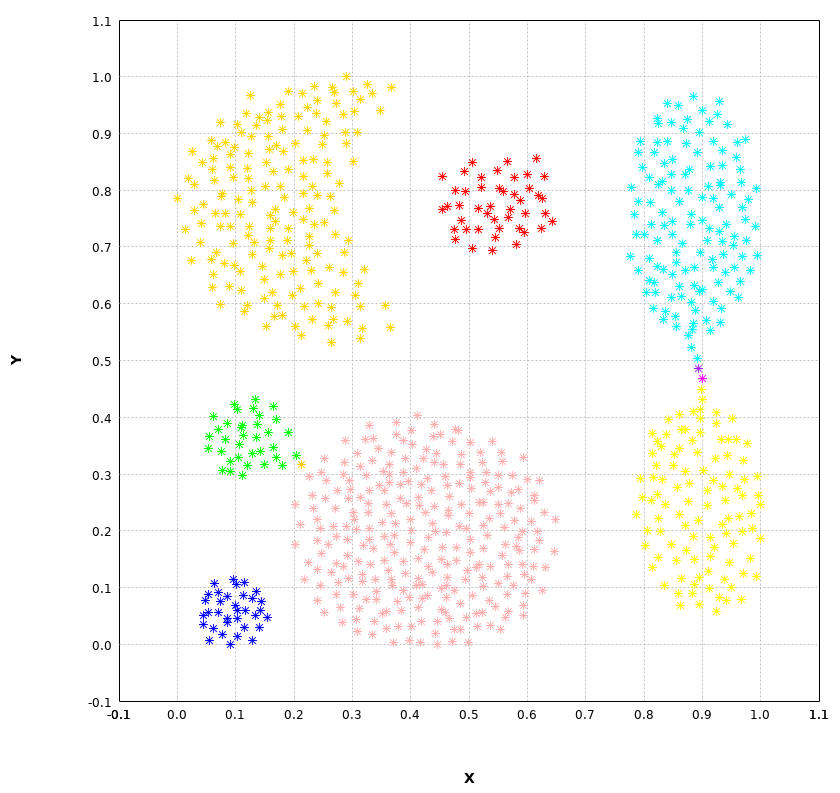}
\caption{NNGA$^{+}$+$\varepsilon$-proximity $[k_1=50]$}
\end{subfigure}
\end{tabular}
\caption{$\varepsilon$-proximity cluster labeling on the Aggregation.
The left image  indicates $\varepsilon$-cluster labeling without NNGA$^{+}$. the right image `NNGA$^{+}$+$\varepsilon$-proximity $[k_1]$' indicates NNGA$^{+}$ with $k_1$ nearest neighbors followed by $\varepsilon$-proximity labeling.}
\label{fig:epsilonClusteringApplications}
\end{figure}

The results for $\varepsilon$-proximity labeling are presented in Figure~\ref{fig:epsilonClusteringApplications}. For the Aggregation dataset, the NNGA$^{+}$ improves the clustering by being able to distinguish between clusters which are joined a filamentary structure. Without NNGA$^{+}$ these filamentary structures cause problems. An analogous improvement in clustering quality is also observed  for the R15 dataset.  The cluster labelling is similar for the Unbalance dataset with  or without NNGA$^{+}$. On the other hand, for the DS1 dataset, the prior application of NNGA$^{+}$ leads to a decrease in the cluster labeling accuracy. 

\subsubsection{Visual evaluation with image segmentation}
\label{image}
A resurgence in interest in the variant of the mean shift algorithm is due to its
application to image segmentation \cite{comaniciu2003} 
where an image is transformed
into a color space in which clusters correspond to segmented regions
in the original image. The 
3-dimensional $L^*u^*v^*$ color space \cite[Eqs.~3.5-8a--f]{pratt2001} is a common choice.
Since an image is a 2-dimensional array of pixels, let $(x,y)$ 
be the row and column index of a pixel.   
The spatial and color (range) information of a pixel can be concatenated into a 
5-dimensional vector $(x,y, L^*, u^*, v^*)$
in the joint spatial-range domain. 
An image segmentation algorithm based on the kernel mean shift was introduced in 
\cite{comaniciu2002} which we adapt for use with NNGA$^{+}$.

\begin{figure}[htbp]
\centering
\setlength{\tabcolsep}{1pt}
\includegraphics[width=0.50\columnwidth]{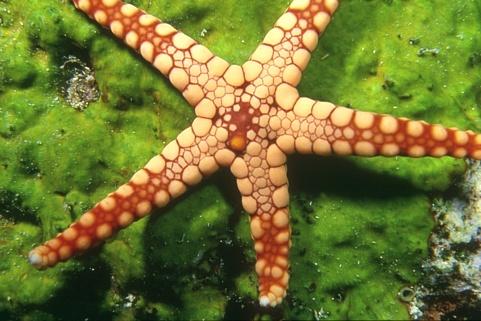}
\caption{Original picture \#36. }
\label{fig:Orig-image-seg}
\end{figure}

Our test image is image \#36 from the colour training set from
the Berkeley Segmentation Dataset and Benchmark\footnote{http://www.eecs.berkeley.edu/Research/Projects/CS/vision/bsds}. Figure~\ref{fig:Orig-image-seg} shows the original RGB 481$\times$321 pixels JPEG image.
The tuning parameters for the NNGA$^{+}$ are $k_1=60$, 
$j_{\max}=15$, 
%
We compute the NNGA$^{+}$-$M_1$ with $M_1=200,400,1000$ buckets.

\begin{figure}[htbp]
\centering
\setlength{\tabcolsep}{1pt}
\begin{tabular}{@{}cc@{}}
(a) NNGA-200  & (b) NNGA$^{+}$-200 (p=1)   \\
\includegraphics[width=0.45\columnwidth]{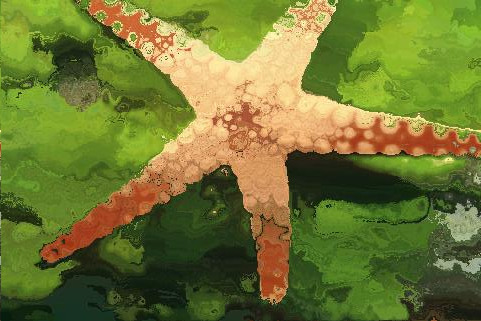} & 
\includegraphics[width=0.45\columnwidth]{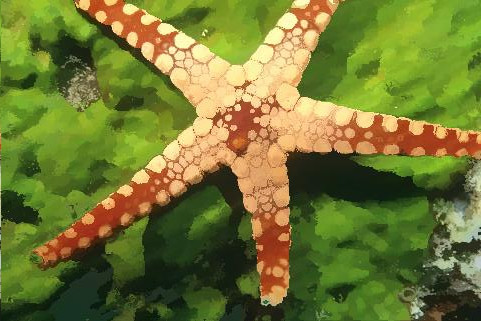}\\
(c) NNGA-400  & (d) NNGA$^{+}$-400(p=1)  \\
\includegraphics[width=0.45\columnwidth]{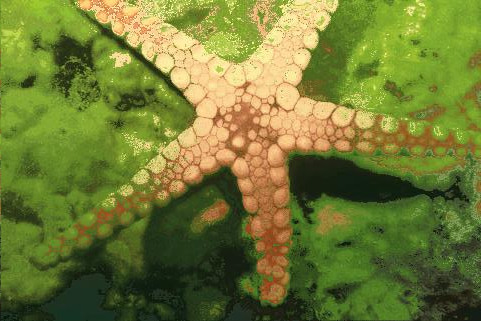} & 
\includegraphics[width=0.45\columnwidth]{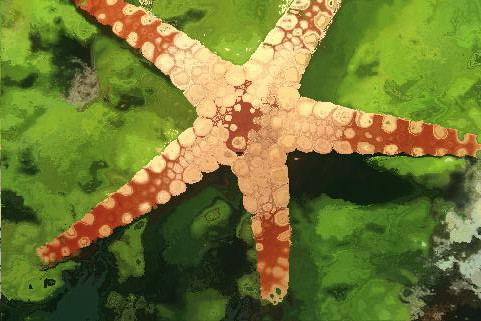} \\
(e) NNGA-1000  & (f) NNGA$^{+}$-1000(p=2)\\
\includegraphics[width=0.45\columnwidth]{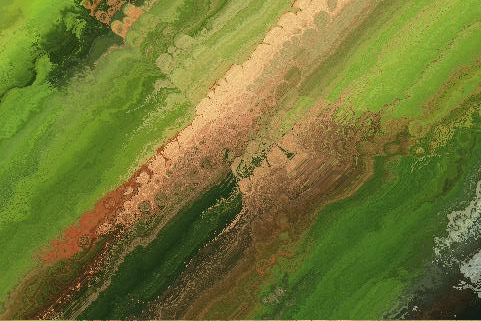} &
\includegraphics[width=0.45\columnwidth]{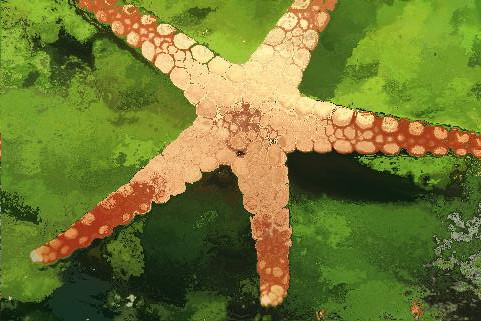}
\end{tabular}
\caption{Colour image segmentation using in the left NNGA and in the right NNGA$^{+}$ with $M_1=200,400$ and 
1000 buckets. }
\label{fig:image-seg}
\end{figure}

For the NNGA$^{+}$-200 and NNGA$^{+}$-400 in Figures~\ref{fig:image-seg}b and \ref{fig:image-seg}d
where we approximate 
nearest neighbours with respectively $M_1=200$ and $M_1=400$ buckets, some finer details are visible, such as the podia.
For NNGA-200 and NNGA-400 in Figures~\ref{fig:image-seg}a and \ref{fig:image-seg}c, the green
background color bleeds into the starfish arms. For NNGA-1000 in Figure~\ref{fig:image-seg}e, the starfish is not visible anymore. However,  by taking more  neighbor layers (NNGA$^{+}$-1000 $p=2$) in Figure \ref{fig:image-seg}f the starfish shape being delimited from the background. 

\begin{figure}[htbp]
\centering
\setlength{\tabcolsep}{1pt}
\begin{tabular}{@{}cc@{}}
(a) User \#{}1109 & (b) User \#{}1119 \\
\includegraphics[width=0.45\columnwidth]{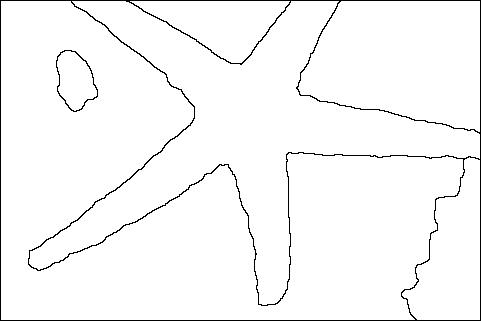} & \includegraphics[width=0.45\columnwidth]{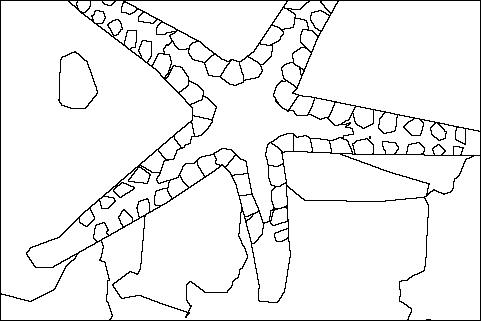} \\
(c) NNGA$^{+}$-200  & (d) NNGA$^{+}$-200 automatic detection \\
\includegraphics[width=0.45\columnwidth]{pictures/img/12003-NNGA-200b-k60-p1.jpg}&
\includegraphics[width=0.45\columnwidth]{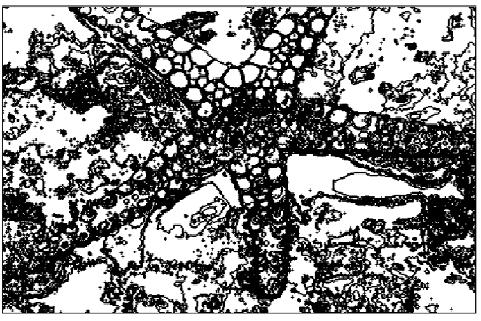} 
\end{tabular}
\caption{Edge detection of segmented images.
(a,b) Two human experts: users \#{}1109 and \#{}1119.  
(c) NNGA$^{+}$-200. (d) NNGA$^{+}$-200 with automatic detection.}
\label{fig:image-edge}
\end{figure}

The Berkeley Segmentation Dataset and Benchmark provides human expert segmentations
of their images for comparisons. 
Figure~\ref{fig:image-edge}a,b depict two edge detections
made by Users \#{}1109 and \#{}1119. User \#{}1109 focuses on segmenting the shape of the starfish, whilst ignoring the detail of the podia, whereas User \#{}1119 concentrates on segmenting the individual podia in the foreground. 
We focus on the NNGA$^{+}$-200 (Figure~\ref{fig:image-seg}b), which
segmentation is closer to User \#{}1119. Whilst this automatic edge detection (Figure~\ref{fig:image-edge}d) in its current state remains too fragmented to be useful for a visual analysis, we anticipate that further application of statistical and image analyses 
will be able to improve it.

\begin{figure*}[htbp]
\centering
\setlength{\tabcolsep}{1pt}
\begin{tabular}{@{}cc@{}}
(a) Picture \#{}91 & (b) NNGA$^{+}$-200 (p=1) \\
\includegraphics[width=0.43\columnwidth]{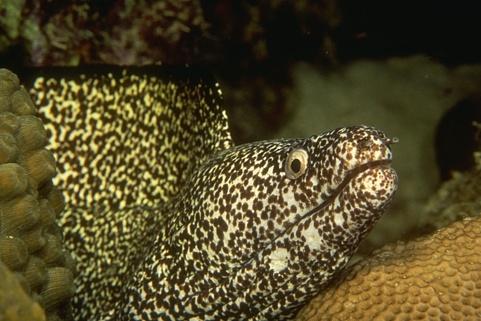} & 
\includegraphics[width=0.43\columnwidth]{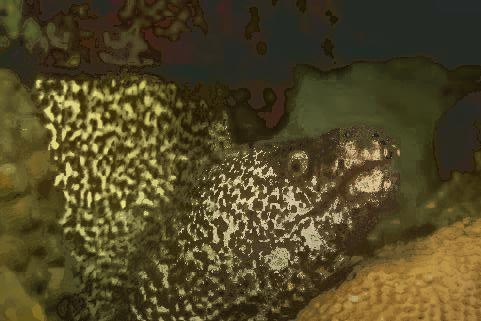} 
\\
(c) Picture \#{}49 & (d) NNGA$^{+}$-200 (p=1) \\
\includegraphics[width=0.43\columnwidth]{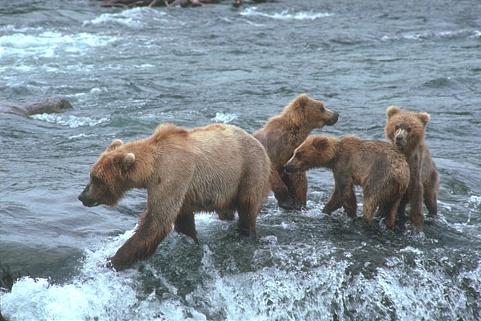}&
\includegraphics[width=0.43\columnwidth]{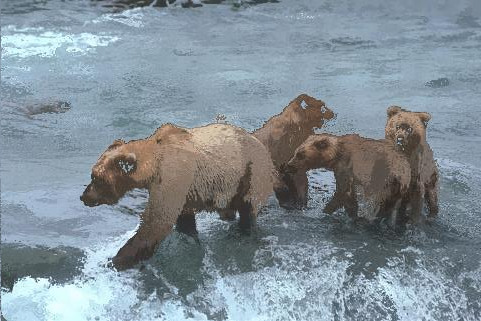}
\\
(e) Picture \#{}81 & (f) NNGA$^{+}$-200 (p=1) \\
\includegraphics[width=0.43\columnwidth]{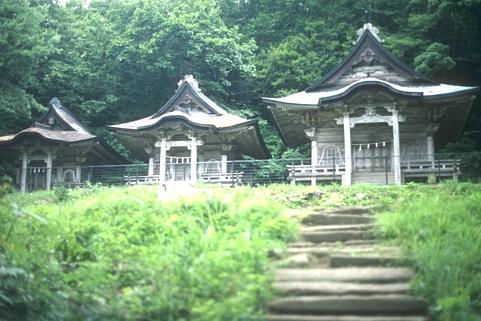}&
\includegraphics[width=0.43\columnwidth]{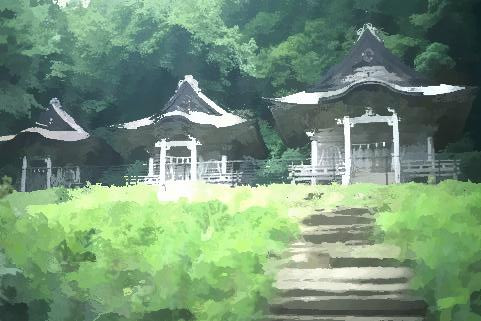} \\
(g) Our picture (5M pixels) & (h) NNGA$^{+}$-5000 (p=1) \\
\includegraphics[width=0.43\columnwidth]{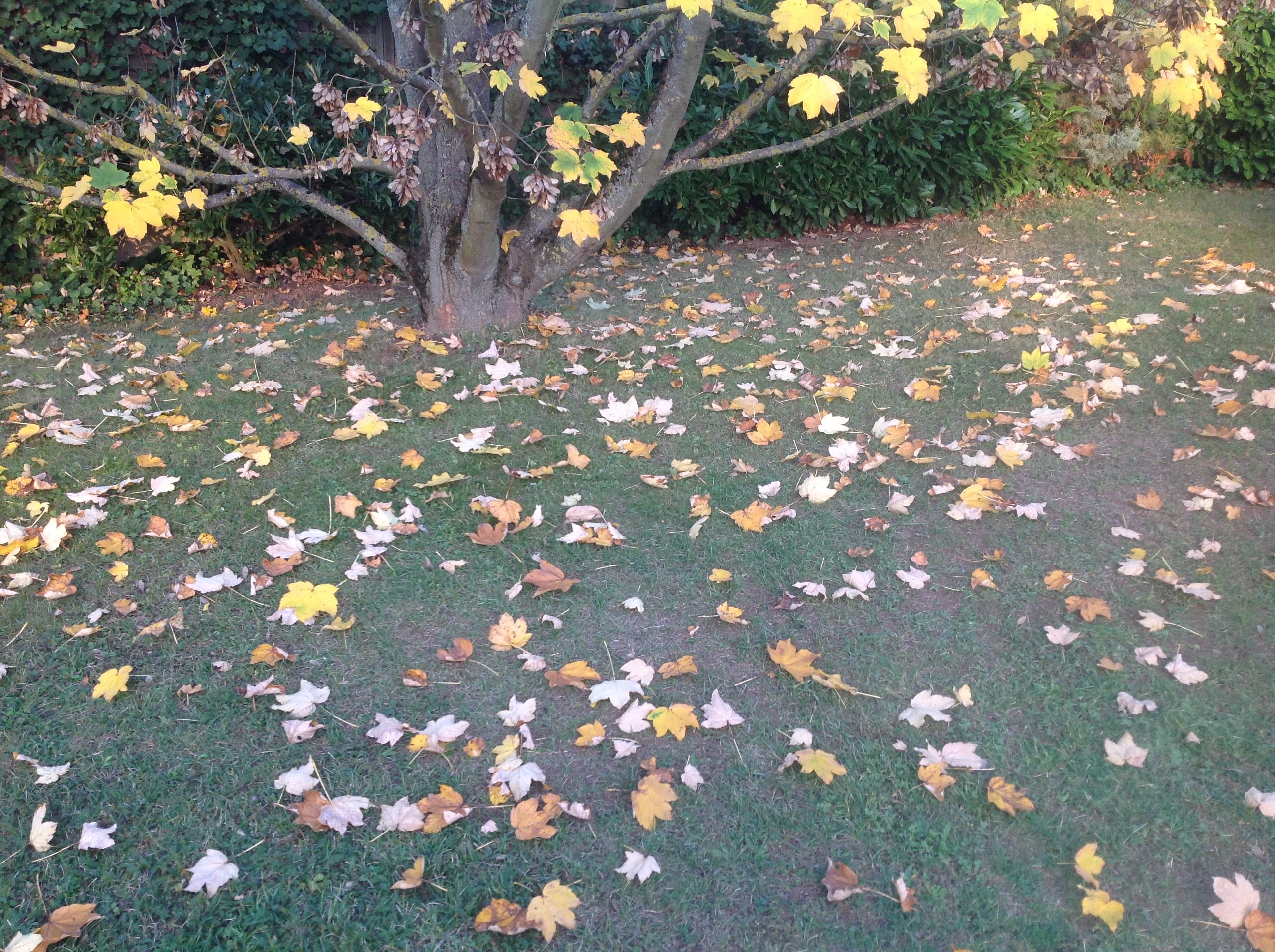} & \includegraphics[width=0.43\columnwidth]{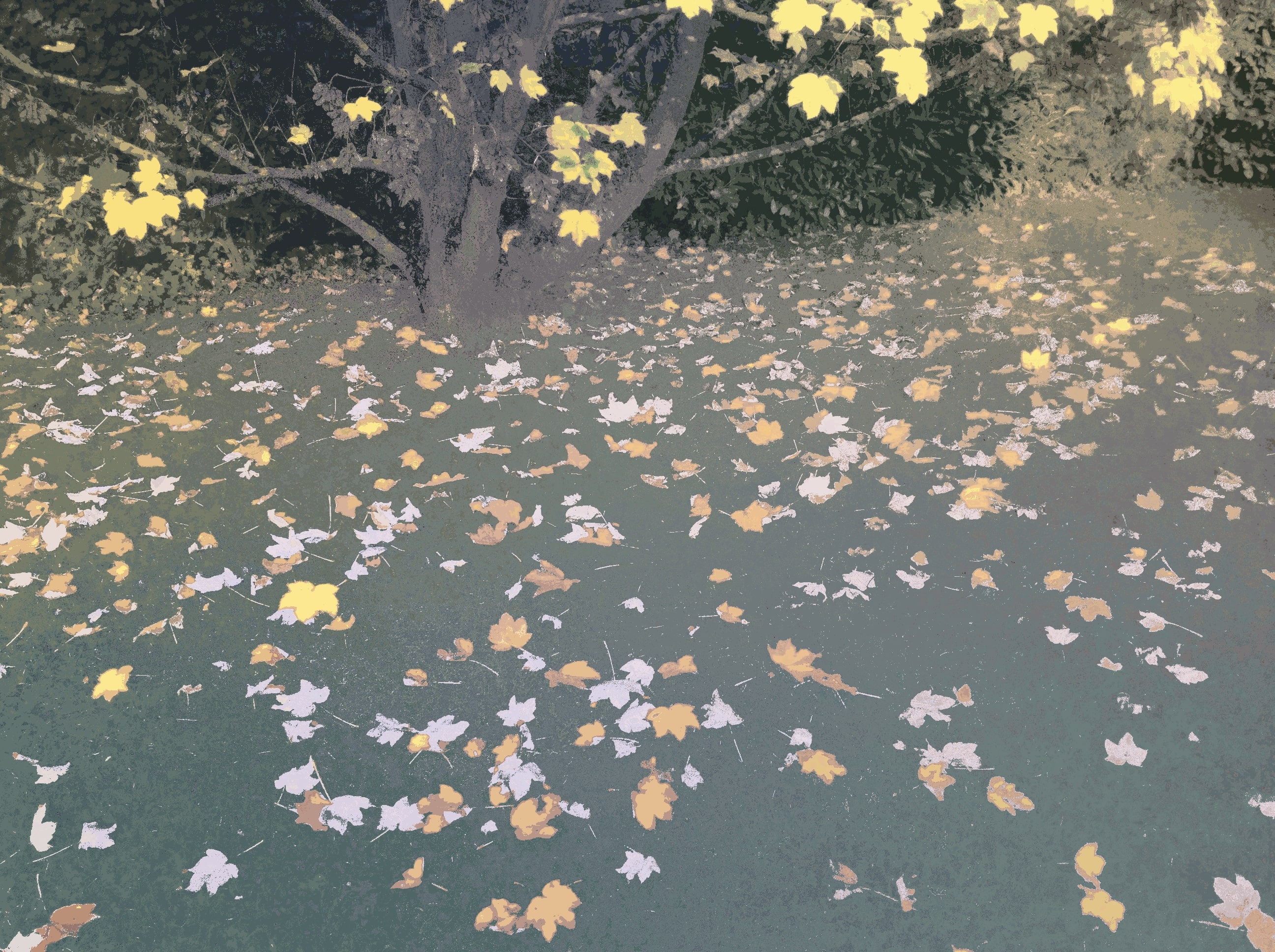} \\
\end{tabular}
\caption{Further examples of segmented images with NNGA$^{+}$-$200$.}
\label{fig:image-example}
\end{figure*}

We observe from Figures~\ref{fig:image-seg} that $M_1=200$ buckets is a suitable  empirical
choice 
and so we apply it to further images from the Berkeley image database as in
Figure~\ref{fig:image-example}. We apply also the NNGA$^{+}$ on
 a picture of 5 millions pixels (taken by ourselves) compared to 150  thousand pixels 
of Berkeley's pictures. 
We increase the number of buckets accordingly 
with the size of the picture to reach approximately 1000 data 
points per bucket. 
The image shows a good 
segmentation between the leafs and the rest of the picture. We distinguish the trunk from the background foliage too without difficulty showing the effectiveness of the algorithm even with big image.

\subsection{Evaluation of scalability}
\begin{figure}[!ht]
\centering
\setlength{\tabcolsep}{1pt}
\begin{tabular}{cc}
\begin{subfigure}[t]{0.5\columnwidth}
\includegraphics[width=\columnwidth]{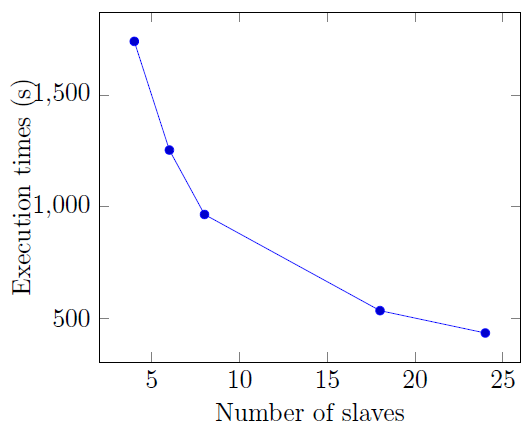}
\caption{ } 
\end{subfigure}
&
\begin{subfigure}[t]{0.5\columnwidth}
\includegraphics[width=\columnwidth]{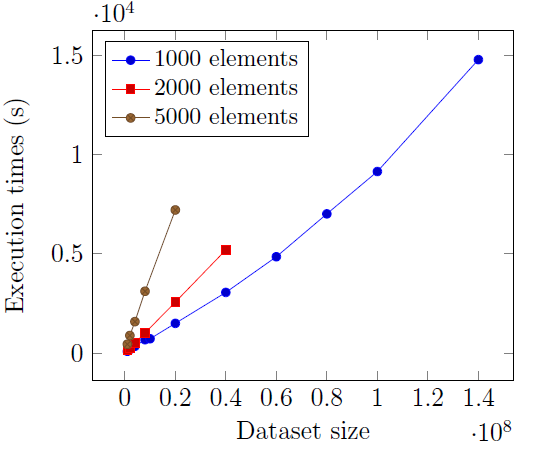}
\caption{ } 
\end{subfigure}
\end{tabular}
\caption{  
(a) Execution times for NNGA$^{+}$ with respect to the number of slaves. 
(b) Execution times for NNGA$^{+}$ with respect to the dataset size $n$ for fixed size buckets.}
\label{fig:GAscalablity}
\end{figure}

\paragraph{Density gradient ascent}
As the data are  distributed over all buckets or Spark partitions, this allows for efficient computations in the local environment of a data point.  Figure~\ref{fig:GAscalablity}a shows that the execution time gradually decreases as the number of slaves increases for a datasets of a fixed size. In Figure~\ref{fig:GAscalablity}b, for a fixed number of nodes, if we maintain the constant number of elements per bucket, the execution time grows linearly with the size of the dataset. This indicates that the number of nearest neighbor $k_1$ needs to be constrained. 

Since we have  an $O((\frac{n}{M_{1}})^2 \log(\frac{n}{M_{1}}))$ complexity per bucket for NNGA$^{+}$, a suitable value for the number of buckets $M_1$ is to keep the $\frac{n}{M_{1}}$ ratio 
approximately equal to  a constant $C$. Then the time complexity of NNGA$^{+}$ reduces to $O(nC\log(C))$. The scalability is demonstrated by the  decrease in execution time with the number of slaves and a linear increase of execution time with the dataset size, reaching  140 million data points which is infeasible for the original quadratic algorithm.

\paragraph{LSH buckets and neighbor layers}

\begin{figure}[!ht]
\centering
\setlength{\tabcolsep}{1pt}
\begin{tabular}{cc}
\begin{subfigure}[t]{0.5\columnwidth}
\includegraphics[width=\columnwidth]{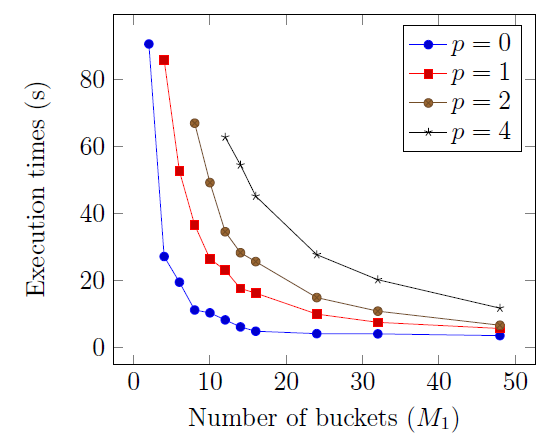}
\caption{ } 
\end{subfigure}
&
\begin{subfigure}[t]{0.5\columnwidth}
\includegraphics[width=\columnwidth]{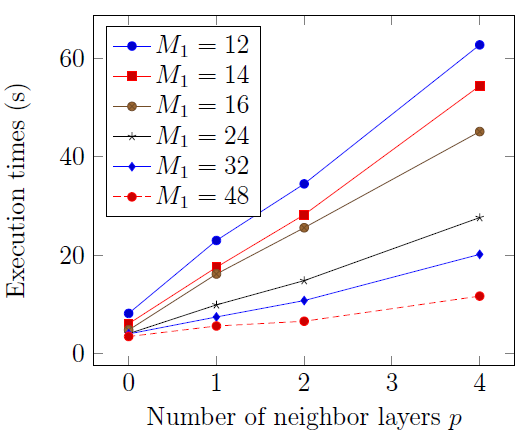}
\caption{ } 
\end{subfigure}
\end{tabular}
\caption{(a) Execution times for NNGA$^{+}$ with respect to the number of buckets ($M_1$). 
(b) Execution times for NNGA$^{+}$ with respect to the number of neighbor layers ($p$).}
\label{fig:nbBucketsGA}
\end{figure}

For a fixed number of neighbor layers, we observe in Figure~\ref{fig:nbBucketsGA}a that the execution time rapidly decreases  and then slows down to reach a plateau, as the number of buckets increases.  
The observed plateau is due to the quadratic complexity of the NNGA$^{+}$: more buckets leads to fewer data points within each bucket and so the execution times can quickly reach the minimal plateau after a sufficiently large number of buckets.
We also studied the influence of the number of neighbors layers $p$ on the execution time.
Whilst NNGA$^{+}$ has quadratic time complexity in each bucket,  if we select an appropriate number of nearest neighbors $k_1$, then we are able to control the execution time of
NNGA$^{+}$ to be linear with respect to the number of neighbors layers $p$, as illustrated in Figure~\ref{fig:nbBucketsGA}b.  

\paragraph{$\varepsilon$-proximity cluster labeling}

\begin{figure}[!ht]
\centering
\setlength{\tabcolsep}{1pt}
\begin{tabular}{cc}
\begin{subfigure}[t]{0.5\columnwidth}
\includegraphics[width=\columnwidth]{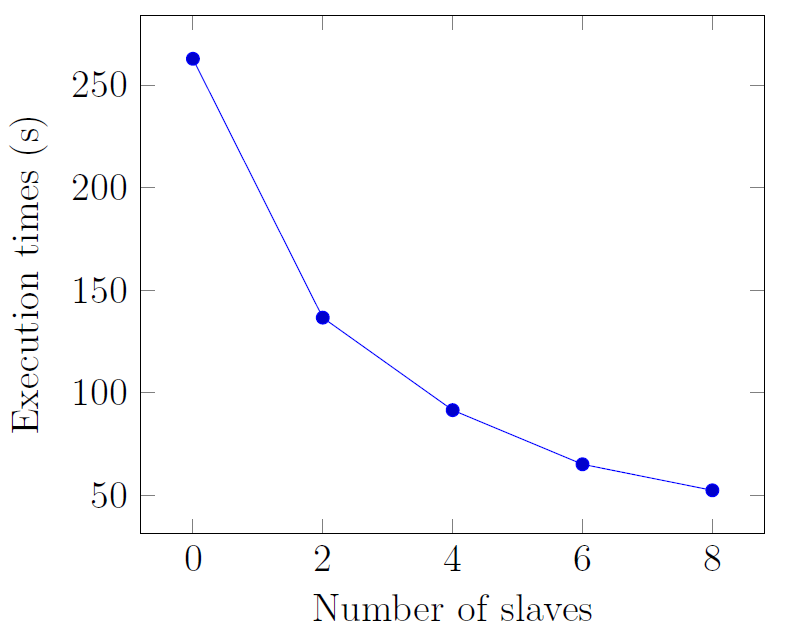}
\caption{ 
}
\label{fig:epsSlavesNb}
\end{subfigure}
&
\begin{subfigure}[t]{0.5\columnwidth}
\includegraphics[width=\columnwidth]{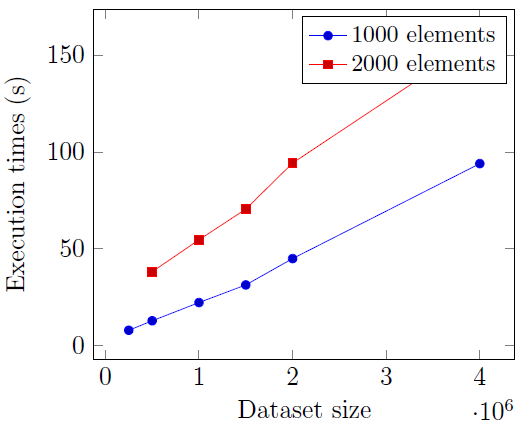}
\caption{
}
\label{fig:epsScalabilityIn}
\end{subfigure}
\end{tabular}
\caption{
(a) Execution times for $\varepsilon$-proximity with respect to the number of slaves.
(b) Execution times for cluster labeling with respect to  the dataset size $n$.
}
\label{fig:epsScalability}
\end{figure}

Concerning $\varepsilon$-proximity cluster labeling,  similar remarks  as for the gradient ascent apply here. As Fig.\ref{fig:epsScalability} follows same decrease in the execution time as for the gradient ascent as a function of the number of slaves and data points we can be confident in the scalability of our approach.

\section{Conclusion}
\label{S:4}
In this paper, we have introduced multiple improvements to the standard nearest neighbors gradient ascent used in Mean Shift algorithm. The first series of improvements are based on new usages of Locality Sensitivity Hashing for approximate nearest neighbors during the nearest neighbors gradient ascent (NNGA$^{+}$), and also during  cluster labeling ($\varepsilon$-proximity). The second one is an efficient and scalable implementation of our ideas on a distributed computing ecosystem based on Spark/Scala. 
We show that using our  NNGA$^{+}$ algorithm, as a pre-processing step in other clustering methods, can  improve  quality   metric evaluations.
We demonstrated that these improvements greatly decrease the execution time whilst maintaining a suitable quality of clustering. These improvements open the opportunity to apply our Mean Shift model for Big Data clustering. 

%
Future work is required to demonstrate the usefullness of NNGA$^{+}$ for clustering algorithms with very high dimensional datasets. Further study on the impact of the chosen dissimilarity measure on the $\varepsilon$-proximity clustering are also required and will be tackled in the future. Optimal choices of the most important tuning parameters for our proposed methods for distributed clustering, namely the number of nearest neighbors for the density gradient ascent, the number of buckets for the LSH, and the threshold for $\varepsilon$-proximity cluster labeling will be a subject of further investigations. Furthermore, we also desire to experiments others hashing techniques \cite{DBLP:journals/corr/WangSSJ14, morvan2018needs} from the litterature in a generic way in order to handle various use case.

\section{Code}
In order to facilitate further experiments and reproducible research, we will provide our contributions through an open source API which will contain NNGA$^{+}$,  $\varepsilon$-proximity algorithms and traditional Mean-shift, $k$-means with Spark/Scala on the following link \url{https://github.com/Clustering4Ever/Clustering4Ever}.
\section*{Acknowledgments}
Experiments presented in this paper were carried out using the Grid'5000 testbed, supported by a scientific interest group hosted by Inria and including CNRS, RENATER and several Universities as well as other organizations (see https://www.grid5000.fr).

\bibliography{biblio}










\end{document}